%% file: main.tex
\documentclass[runningheads]{llncs}

\usepackage{eccv}

\usepackage{eccvabbrv}

\usepackage{graphicx}
\usepackage{booktabs}

\usepackage[accsupp]{axessibility}  

\usepackage{hyperref}

\usepackage{orcidlink}

\usepackage[ruled,vlined,linesnumbered]{algorithm2e}
\usepackage{booktabs}
\usepackage{amssymb}
\usepackage{pifont}
\usepackage{xcolor}
\usepackage{colortbl}
\usepackage{booktabs}
\usepackage{makecell}
\usepackage[table]{xcolor}
\usepackage{adjustbox}
\usepackage{comment}
\usepackage{mathtools} 
\usepackage{multirow}

\usepackage{listings}
\usepackage{xcolor}

\usepackage{enumitem}
\usepackage{parskip}
\makeatletter
\renewcommand\section{\@startsection{section}{1}{\z@}%
                       {-18\p@ \@plus -4\p@ \@minus -4\p@}
                       {12\p@ \@plus 4\p@ \@minus 4\p@}%
                       {\normalfont\large\bfseries\boldmath
                        \rightskip=\z@ \@plus 8em\pretolerance=10000 }}
\renewcommand\subsection{\@startsection{subsection}{2}{\z@}%
                       {-16\p@ \@plus -4\p@ \@minus -4\p@}%
                       {8\p@ \@plus 4\p@ \@minus 4\p@}%
                       {\normalfont\normalsize\bfseries\boldmath
                        \rightskip=\z@ \@plus 8em\pretolerance=10000 }}
\makeatother

\definecolor{task_color}{RGB}{48,115,81}
\definecolor{domain_color}{RGB}{27,64,121}
\definecolor{attn_color}{RGB}{163,22,33}

\definecolor{darkgreen}{RGB}{0,120,0}
\definecolor{darkred}{RGB}{160,0,0}

\definecolor{hlrow}{RGB}{235,245,255}

\newcommand{\hec}[1]{\cellcolor{black!6}#1}

\newcommand{\TODO}[1]{{\color{red}TODO #1}}
\newcommand{\ourmethod}{HEC\xspace}

\begin{document}

\title{%
Unlocking Few-Shot Capabilities in LVLMs via Prompt Conditioning and Head Selection} 

\titlerunning{HEC}

\author{Adhemar de Senneville\inst{1}\orcidlink{0009-0004-6362-6358} \and
Xavier Bou\inst{2}\orcidlink{0000-0002-1799-9506} \and
Jérémy Anger\inst{1}\orcidlink{0009-0007-3319-6037} \and
Rafael Grompone\inst{1}\orcidlink{0000-0002-6309-7116} \and
Gabriele Facciolo\inst{1,3}\orcidlink{0000-0002-8855-8513}}

\authorrunning{A. de Senneville et al.}

\institute{Université Paris-Saclay, CNRS, ENS Paris-Saclay, Centre Borelli, France \and
École Polytechnique, AMIAD, France \and 
Institut Universitaire de France \\
\email{adhemar.de\_senneville@ens-paris-saclay.fr}}

\maketitle
\input{sec/1_intro}

\input{sec/2_method}
\input{sec/3_exp}

\section*{Acknowledgements}
This work was partially funded by AID-DGA (l’Agence de l’Innovation de D\'{e}fense a la Direction G\'{e}n\'{e}rale de l’Armement, Minit\`{e}re des Armees), and was also partly funded by the ANR-DFG project BOFOR ANR-24-CE92-0048.
This work was granted access to the HPC resources of IDRIS under the allocations 2025-AD011016525  made by GENCI.     


%
%
\bibliographystyle{splncs04}
\bibliography{main}

\clearpage
\input{sec/4_supp}

\end{document}

%% file: sec/1_intro.tex

\begin{abstract}
Current Large Vision Language Models (LVLMs) excel at 
many zero-shot tasks like image captioning, visual question answering and OCR.
However, these same models suffer from poor performance at image classification tasks, underperforming against CLIP-based methods. 
Notably, this gap is surprising because many LVLMs use CLIP-pretrained vision encoders.
Yet LVLMs are not inherently limited by CLIP’s architecture with independent vision and text encoders.
In CLIP, this separation biases classification toward class-name matching rather than joint visual–text reasoning.
In this paper we show that, despite their poor raw performance, 
LVLMs can improve visual feature class separability at inference using prompt conditioning, and 
LVLMs’ internal representations, especially attention heads, can outperform the model itself at zero-shot and few-shot classification.
We introduce Head Ensemble Classifiers (HEC) to
bridge the performance gap between CLIP-based and LVLM-based classification methods.
Inspired by Gaussian Discriminant Analysis, HEC ranks the most discriminative vision and text heads and combines them into a training-free classifier.
We show that HEC achieves state-of-the-art performance in
few-shot and zero-shot classification across 12 datasets.
Code: \href{https://github.com/AdhemarDeSenneville/HEC}{\texttt{github.com/AdhemarDeSenneville/HEC}}
  \keywords{Zero shot classification \and Few shot classification \and Large Vision-Language Model \and CLIP }
\end{abstract}


\begin{figure}[tb]
  \centering
  \includegraphics[height=5.3cm]{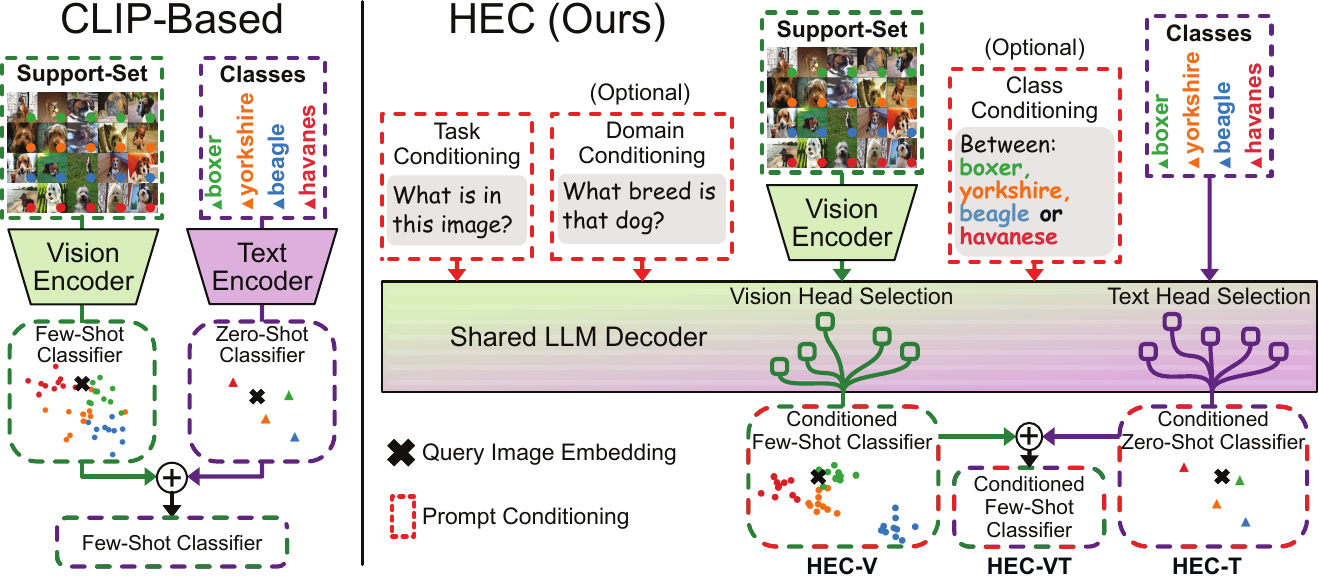}
  \caption{
  \textbf{CLIP-based vs HEC (Ours).} 
  CLIP-based methods encode class names and support set images independently to construct a zero-shot and a few-shot classifier respectively. 
  Our method keeps the same two-classifier structure.
  However, both distributions go through a shared LLM decoder which can be conditioned by a text prompt to include guidance on the domain or the classes of the support set.
  The few-shot classifier (HEC-V) builds on the distribution of a sparse set of heads from the LLM decoder. 
  This subset, which we refer to as \textit{vision-heads}, is selected using Gaussian Discriminant Analysis~\cite{bishop2006prml} (\cref{fig:main})%
  .
  The zero-shot classifier (HEC-T) builds on the distribution of another sparse set of heads, which we refer to as \textit{text-heads}.
  Similarly to CLIP-based methods, the two classifiers can be combined in a single one (HEC-VT) by adding their output probabilities.
  }
  \label{fig:comparison}
\end{figure}

\section{Introduction}
\label{sec:intro}

Recent LVLMs 
show remarkable progress in a wide range of computer vision tasks, including image captioning~\cite{chen2024lion,li2023blip}, text transcription~\cite{Liao_2025_CVPR_ocr}, Visual Question Answering (VQA)~\cite{NEURIPS2023_6dcf277e} and grounding~\cite{Ferret,peng2023kosmos}. 
In addition, they offer the versatility to address all of these problems with a single pre-trained model and no additional fine-tuning.
However, LVLMs still lag behind the state of the art in few-shot image classification~\cite{liu2024improved_whybad, whybad}, particularly compared to CLIP-based models~\cite{CLIP_CLIP,CLIP_Proker,CLIP_TIMO,CLIP_GDA}.
This is surprising as many LVLMs inherit from a CLIP~\cite{CLIP_CLIP} vision encoder, yet score below CLIP at zero-shot~\cite{whybad}.

Nonetheless, despite exhibiting strong performances, CLIP-based methods still have some limitations.
First, CLIP mainly matches images to class names or descriptions, making it weaker when depending on domain text contextualization~\cite{qu2025proapo,fahes2022p,cao2024domain}. 
Secondly, CLIP vision and text encoders are independent, so image--text interaction is limited at decision time, as illustrated in Fig.~\ref{fig:comparison} (left). 
In contrast, LVLMs use a vision encoder to convert an image into a sequence of vision tokens, or use a tokenizer to convert a text prompt into a sequence of text tokens.
Then, these vision and text tokens are jointly processed by a shared LLM transformer decoder allowing image--text interactions during inference.

There is a mismatch between the rich internal representations that LVLMs should inherit from their CLIP encoder and their weak final output predictions~\cite{whybad}.
Consequently, previous works either focus on
using LVLMs to generate captions to improve CLIP performance \cite{NIPS_vlm,AAAI_vlm},
or extensive finetuning on  downstream classification tasks \cite{Finedefics,ouali2025vladva_cliplike}.
This mismatch led us to investigate whether the LVLMs’ internal representations can be leveraged for few-shot classification.

Inspired by Gaussian Discriminant Analysis (GDA)~\cite{bishop2006prml,CLIP_GDA} and recent works on training-free LVLM adaptation~\cite{SAV,MTV},
we propose two simple yet 
effective mechanisms to condition and extract LVLMs multimodal representations. 
First, we use text prompts to condition LVLMs feature distribution during inference. The prompt, concatenated with vision tokens, specifies contextual information of the few-shot task, as shown in Fig.~\ref{fig:comparison}.
Then, we extract features by selecting a sparse set of 
attention heads that maximize few-shot and zero-shot classification performance.
More specifically we select the best attention heads at few-shot (denoted \textit{vision-heads}) using GDA, selecting heads that maximize performance given a set of labeled images (support set).
Similarly, we introduce a mechanism to select the best attention heads at zero-shot (denoted \textit{text-heads}). 
 
Building on these two mechanisms, we introduce three distinct Head Ensemble Classifiers (HEC), see \cref{fig:comparison}.
When class names are unknown, we show that combining vision-heads produces a few-shot classifier (HEC-V) that improves its performance by conditioning the classification with a domain-specific prompt (e.g., \texttt{What breed is that dog?}). This type of prompt guidance domain adaptation is not possible with vision models and CLIP-based models.  %
We also show that combining text-heads into a classifier (HEC-T) improves for free the zero-shot accuracy of a given LVLM.
Lastly, we bridge the performance gap between training-free CLIP-based methods and LVLM-based methods by introducing a classifier that combines text-heads and vision-heads (HEC-VT).

In summary, our contributions are:
\begin{itemize}
  \item We experimentally show that adding prompt conditioning to LVLMs improves the class separability of its internal distributions, especially within a sparse set of heads.
  \item We introduce a method to select and combine those top heads at test-time.
  \item We report state-of-the-art performance at few-shot and zero-shot classification, bridging the gap between CLIP-based and LVLM-based methods without the need of additional fine-tuning.
\end{itemize}


%
%

\begin{figure}[tb]
  \centering
  \includegraphics[height=4.5cm]{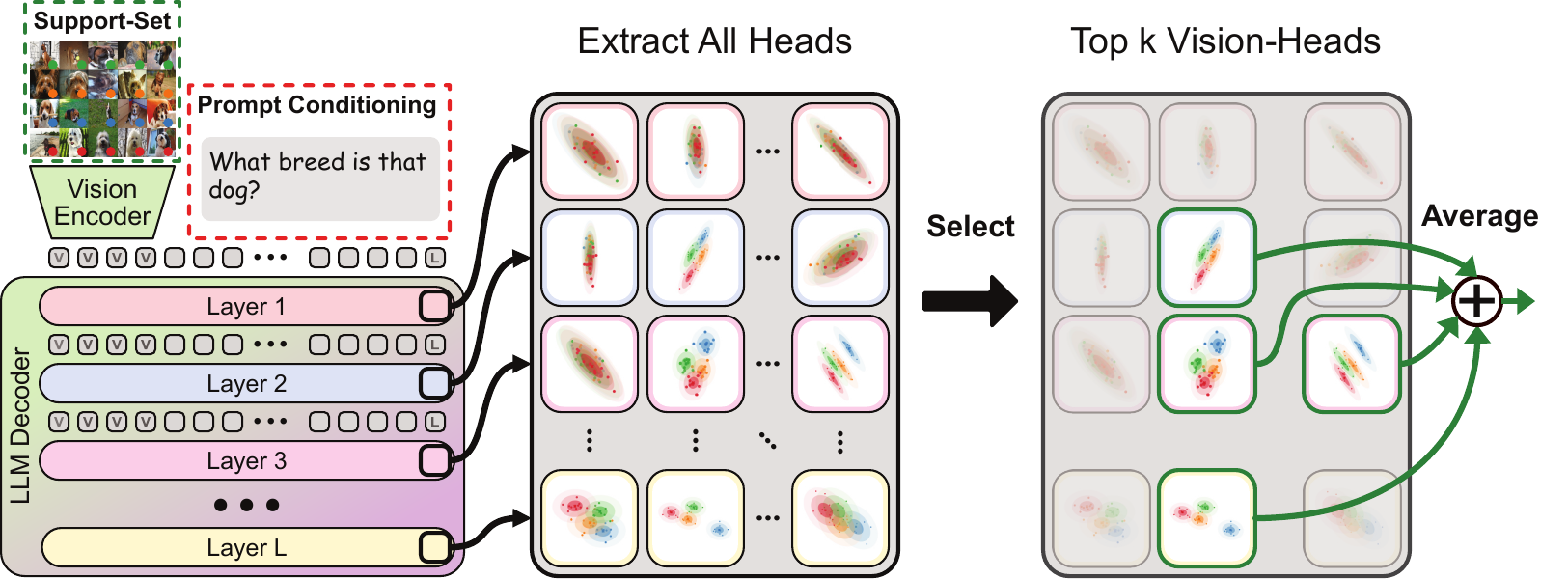}
  \caption{
  \textbf{Overview of HEC-V.} 
  Given a prompt, we first encode all the images from the support set with the LVLM.
  We then extract the distribution of attention vectors \eqref{eq:head_extract} for the last token across all heads in every layer.
  Then, based on a Gaussian Discriminant Analysis~\cite{bishop2006prml}, we rank each head based on its class separability.
  Lastly, given a query image, we ensemble the predictions of the top $k$ heads for that task by averaging their class probabilities.}
  \label{fig:main}
\end{figure}


\section{Previous Work}

\paragraph{Training-Free CLIP}
Most research effort around multimodal few-shot image classification has been focused around CLIP-based architectures. 
CLIP in its original formulation can be applied easily to zero-shot image classification~\cite{CLIP_CLIP} using the known class names. 
Prior work shows that CLIP’s zero-shot classification can be improved with parameter-free attention maps~\cite{CLIP_CALIP}, similarly we propose a training-free adaptation of LVLMs for zero-shot classification.
On the other hand, when class names are unknown but a small labeled support set is available, the training-free baselines include nearest-centroid classifier~\cite{snell2017prototypical,chen2021meta} and linear probing with closed-form solutions~\cite{bertinetto2018meta, CLIP_GDA, balestriero2025lejepa}. %
When both class names and the support set are available, a first line of work simply adds logits from a zero-shot and a few-shot classifier together~\cite{CLIP_tipadapter,CLIP_GDA}.
Instead, some improve upon this by combining text and visual frozen features to build a single classifier~\cite{CLIP_Proker,CLIP_TIMO,CLIP_APE}.



\vspace{-.5em}
\paragraph{LVLMs in Few-Shot Learning} 
A first application of LVLMs to few-shot learning was to use them to generate descriptions to guide a CLIP-based classifier~\cite{NIPS_vlm,AAAI_vlm,liu2024democratizing_vlm}, rather than using the LVLM as the classifier.
To mitigate low classification performance, a line of work directly fine-tunes the model on fine-grained classification tasks.
Finedefics~\cite{Finedefics} is trained using attribute descriptions and CLIP-like contrastive losses.
In~\cite{liu2025making}, meta-training improves in-context learning performance.
Recent work explicitly fine-tunes LVLMs as CLIP encoders~\cite{ouali2025vladva_cliplike}, training them with contrastive objectives aligning image and text embeddings~\cite{yu2025cafe_cliplike,jiangvlm2vec_cliplike}.
Similarly to us, 
VLM2Vec~\cite{jiangvlm2vec_cliplike} uses instruction conditioning 
(e.g. \texttt{Instruction: Represent the given image and the related question}), 
which we refer to as Task conditioning. 
However, in our work, performance gains arise from adding domain and class conditioning.
Our method can be applied to any LVLM, making it complementary to fine-tuning approaches.

\vspace{-.5em}
\paragraph{Training-Free LVLM}
A straightforward training-free method to improve performance is in-context learning, where few example images per class are added to the prompt~\cite{alayrac2022flamingo}. 
However, the performance decreases rapidly with the number of shots and classes present in the prompt~\cite{chen2025mmices,huang2025mimicking,santos2025vision_incontext}.
Some lines of work mitigates this by introducing task vectors that compress many in-context tasks in a single prompt~\cite{hendel2023context,hojel2024finding}.
MTV~\cite{MTV} improved on that furthermore by selecting task vectors inside a sparse set of attention heads.
Recently SAVs~\cite{SAV} 
directly selects top heads from the support set using nearest centroid classifier without having to compute task vectors.
Our work builds on SAVs by improving the head selection and ensemble mechanisms.

%% file: sec/2_method.tex
\section{Preliminaries}
\label{sec:preliminaries}

In this section, we first formalize the few-shot and zero-shot image classification setup. 
We then conduct a preliminary investigation 
on where LVLMs build their representation during an image classification task. 
Particularly we observe that (1) prompt conditioning allows the last token of the LLM decoder to build multi-modal representations that outperform the vision backbone. (2) A sparse set of attention heads contain representations that outperform the prediction based on the last token. 


\subsection{Problem Formulation}

We consider the $N$-way few-shot image classification over classes $\mathcal{C}=\{1,\dots,N\}$. The support set is composed of image--label pairs \(\{(x_i,y_i)\}_{i=1}^{NK}\), with only a small number of \(K\) samples per class ($K$-shots). 
A class text $t_c$ is associated to each label. 
The query set is the set of unlabeled images we aim to classify.

In the training-free paradigm, we use a frozen foundation model to represent both images and class texts in a shared embedding space \cite{CLIP_tipadapter,CLIP_APE,CLIP_GDA}. 
We denote $z_i^{(\text{v})}$ the embedding of a support image $x_i$, and $z_c^{(\text{t})}$ the embedding of the class text $t_c$.
In the \textbf{text-zero-shot} setting, the class logits are given by the dot product $z_q^{(\text{v})\top}z_c^{(\text{t})}$ where $z_q^{(\text{v})}$ is the query image embedding.
In the \textbf{vision-few-shot} setting, we define a classifier from the support set embeddings $\{z_i^{(\text{v})}, y_i\}_{i=1}^{NK}$, to predict the class of the query embedding $z_q^{(\text{v})}$.
In the \textbf{vision-text-few-shot} setting, we are given $\{z_c^{(\text{t})}\}_{c\in\mathcal{C}}$ and $\{z_i^{(\text{v})}, y_i\}_{i=1}^{NK}$. A simple option to address this problem is to add the logits of the text-zero-shot and a vision-few-shot classifiers \cite{CLIP_tipadapter,CLIP_GDA}.
However, some methods introduced vision--text coupling by treating the text-zero-shot classifier as a fixed prior and adjusting its logits using the support set \cite{CLIP_APE,CLIP_TIMO,CLIP_Proker}.


\vspace{-.5em}\paragraph{CLIP encoder:}
Extracting text and image CLIP embeddings is straightforward. It uses two separate encoders, the vision encoder \(z_i^{(\text{v})} = f^{v}(x_i)[\text{CLS}]\) and the text encoder \(z_c^{(\text{t})} = f^{t}(\texttt{"a photo of a \{$t_c$\}"})[\text{CLS}]\), to map images and class prompts into a shared embedding space. Here, $[\text{CLS}]$ denotes the extraction of the class token, meaning that the embedding is  a single token.

\vspace{-.5em}\paragraph{LVLM as a CLIP encoder:}
Unlike CLIP-based methods, encoding text and images using LVLMs is not straightforward.
Inspired by \cite{ouali2025vladva_cliplike,yu2025cafe_cliplike,jiangvlm2vec_cliplike}, we propose a framework for using LVLMs as CLIP encoders.
Let $f^{v}$ be the vision encoder that maps an image $x_i$ to a sequence of vision tokens, and let $\text{LLM}(\cdot)$ be the LLM decoder that takes that sequence of vision tokens as well as text tokens as input.
For instance, Qwen2-VL has a ViT visual encoder with 32 layers and an LLM decoder with 28 layers.
We extract the output embedding from the summary token, i.e.  $\mathrm{ST}(\cdot)$, which returns the last token of the last layer embedding of the LLM decoder~\cite{ouali2025vladva_cliplike}. 
Unlike CLIP, LVLM embeddings can be prompt-conditioned as the LLM decoder jointly processes vision tokens with text tokens from the prompt $\pi$.
The LLM decoder input is the concatenation $[f^{v}(x_i);\pi]$ (see Fig.~\ref{fig:main}), and the embedding of an image is obtained as
\begin{equation}
z_i^{(\text{v})}=\text{ST}(\text{LLM}([f^{v}(x_i);\pi])).
\label{eq:lvlm_encode_vision}
\end{equation}
To obtain the class embedding, we replace image tokens by a textual class description: 
\begin{equation}
z_c^{(\text{t})}=\text{ST}(\text{LLM}([\texttt{"You are given an image of a \{$t_c$\}."};\pi])).
\label{eq:lvlm_encode_text}
\end{equation}
The prompt $\pi$ can incorporate incrementally different levels of guidance depending on available information. Thus, we propose three levels of conditioning:
\begin{itemize}
  \item\textbf{\textcolor{task_color}{Task Conditioning}}: A task-specific prompt (e.g., \texttt{What is the object in the image?}) 
  pushes the summary token toward discriminative representation.
  As shown in \cite{ouali2025vladva_cliplike}, adding constraints in the prompt such as \texttt{Answer in one word} can improve performance by encouraging the model to compress information in the next-token representation.
  It is important to note that, at this stage, by switching the prompt we can solve other classification tasks 
  such as VQA, image--text pair classification or image retrieval~\cite{SAV}.
  \item\textbf{\textcolor{domain_color}{Domain Conditioning}}: Similarly to the case of task guidance, in fine-grained settings, rephrasing the prompt to be domain-specific (e.g., \texttt{What breed is that
dog?}) should additionally push the summary token toward domain-discriminative representation.
  \item\textbf{\textcolor{attn_color}{Class Conditioning}}: Moreover, if the candidate classes are known, appending to the domain prompt the class list (e.g., \texttt{Between: boxer, yorkshire, beagle or havanese.}) should additionally push the summary token toward class-discriminative representation.
\end{itemize}

\vspace{-.5em}\paragraph{Head Extraction}
In the following we formalize how features from a given head are extracted.
We index attention heads by \(m \in \{1,\dots,M\}\), where the total number of heads is 
$M=L \cdot H$ with $L$ the number of layers, and $H$ the number of heads per layer.
For each head $m$, we compute the corresponding head embedding as
\begin{equation}
\mathbf h_{m}
=
\operatorname{softmax}\!\left(\frac{\mathbf q_m\,\mathbf K_m^\top}{\sqrt{D}}\right)\mathbf V_m,
\label{eq:head_extract}
\end{equation}
where $\mathbf q_m$ is the query vector of only the last token in the input sequence, $\mathbf K_m$ and $\mathbf V_m$ are the key and value matrices of the current head at the current layer, and \(D\) is the head dimension.
For the rest of the method, we L2-normalize each $\mathbf h_{m}$ making dot products equivalent to cosine similarities.
Following~\cite{SAV} we denote $\mathbf h_{m}$ as an \textit{attention vector} for head $m$.
We denote $\mathbf h^{(\text{v})}_{i,m}$ the attention vector when encoding the image $x_i$ and 
$\mathbf h^{(\mathrm{t})}_{c,m}$ when encoding the text class $t_c$.

\subsection{What happens inside LVLMs} 

\begin{figure}[t]
    \centering
    \begin{subfigure}{0.38\linewidth}
        \centering
        \includegraphics[width=\linewidth]{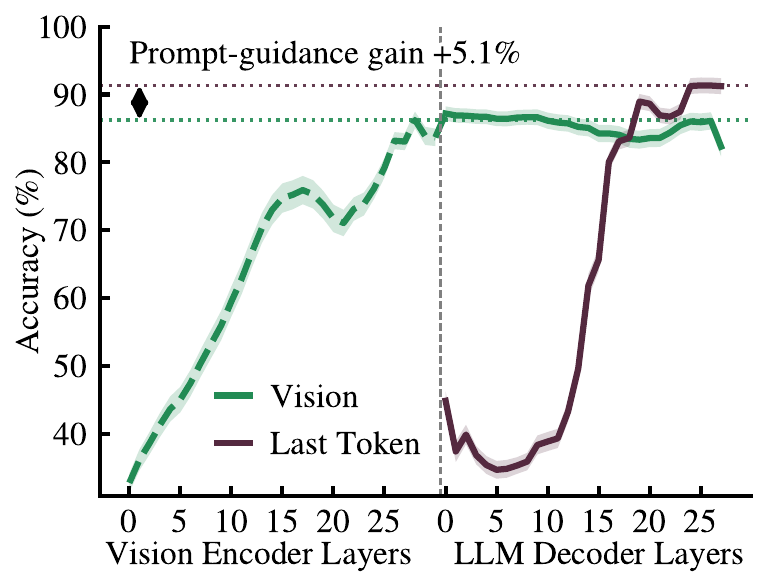}
        \captionsetup{width=.92\linewidth}
        \caption{Evolution of embedding few-shot accuracy across vision encoder layers and LLM decoder layers}
        \label{fig:qwen_probing_a}
    \end{subfigure}\hfill
    \begin{subfigure}{0.30\linewidth}
        \centering
        \includegraphics[width=\linewidth]{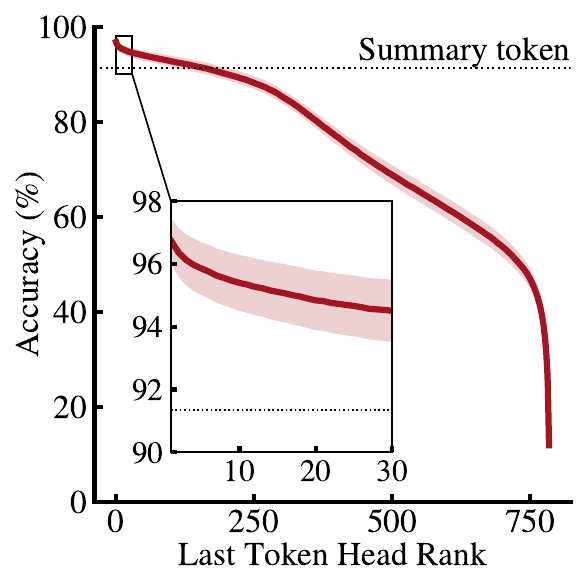}
        \captionsetup{width=.92\linewidth}
        \caption{Few-shot accuracy of all last token heads, ranked from best to worst}
        \label{fig:qwen_probing_c}
    \end{subfigure}
    \begin{subfigure}{0.30\linewidth}
        \centering
        \includegraphics[width=\linewidth]{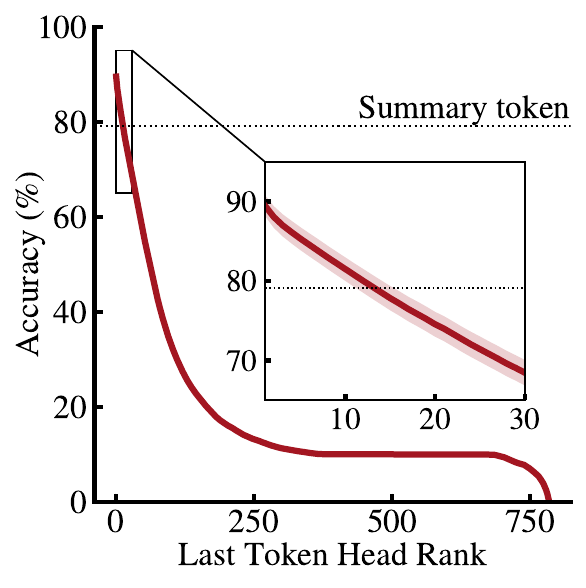}
        \captionsetup{width=.92\linewidth}
        \caption{Zero-shot accuracy of all last token heads, ranked from best to worst}
        \label{fig:qwen_probing_d}
    \end{subfigure}
    \vspace{-.5em}
    
    \caption{
    Experiments to identify where the best classification representation lies in LVLMs.
    At different locations of Qwen2-VL, we compute linear probing accuracy averaged over a thousand 10-way 4-shot tasks across 10 datasets using \textcolor{attn_color}{Class} conditioning.
    (a) Although vision tokens yield strong accuracy early on, inherited from the CLIP vision transformer, the last token builds better representations by integrating multimodal features from vision and text prompt tokens.
    (b) For each few-shot setup, a small number of top heads called \textit{vision-heads} yield better performance than the summary token.
    (c) Similarly, for each zero-shot setup, a small number of top heads called \textit{text-heads} yield better performance than the summary token.
    }
    \label{fig:qwen_probing_row}
\end{figure}

\begin{figure}[t]
  \centering
  \setlength{\tabcolsep}{6pt} 
  \renewcommand{\arraystretch}{1.0}
  \begin{tabular}{@{}c c@{}}
    \includegraphics[width=0.35\textwidth]{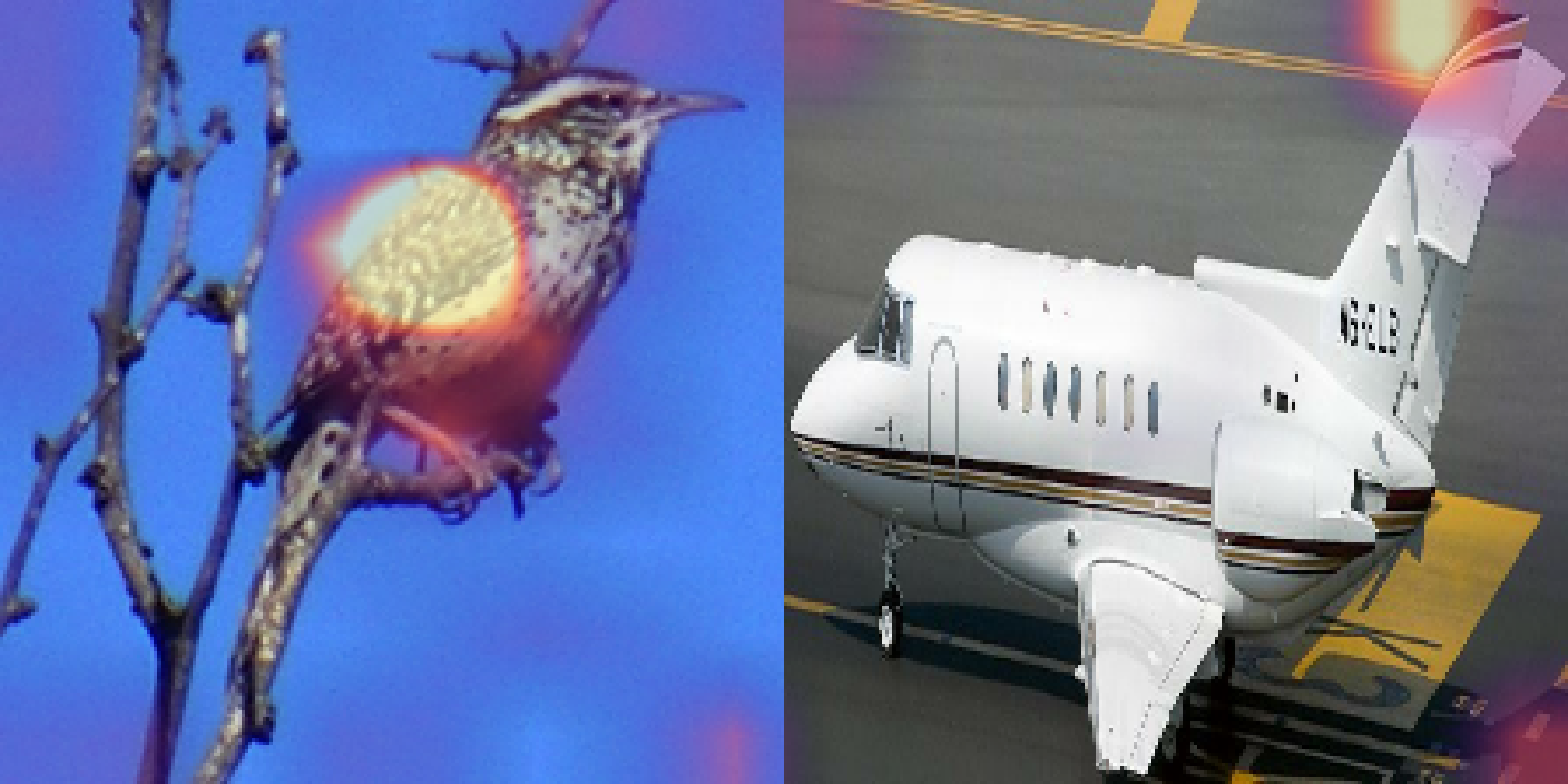} &
    \includegraphics[width=0.35\textwidth]{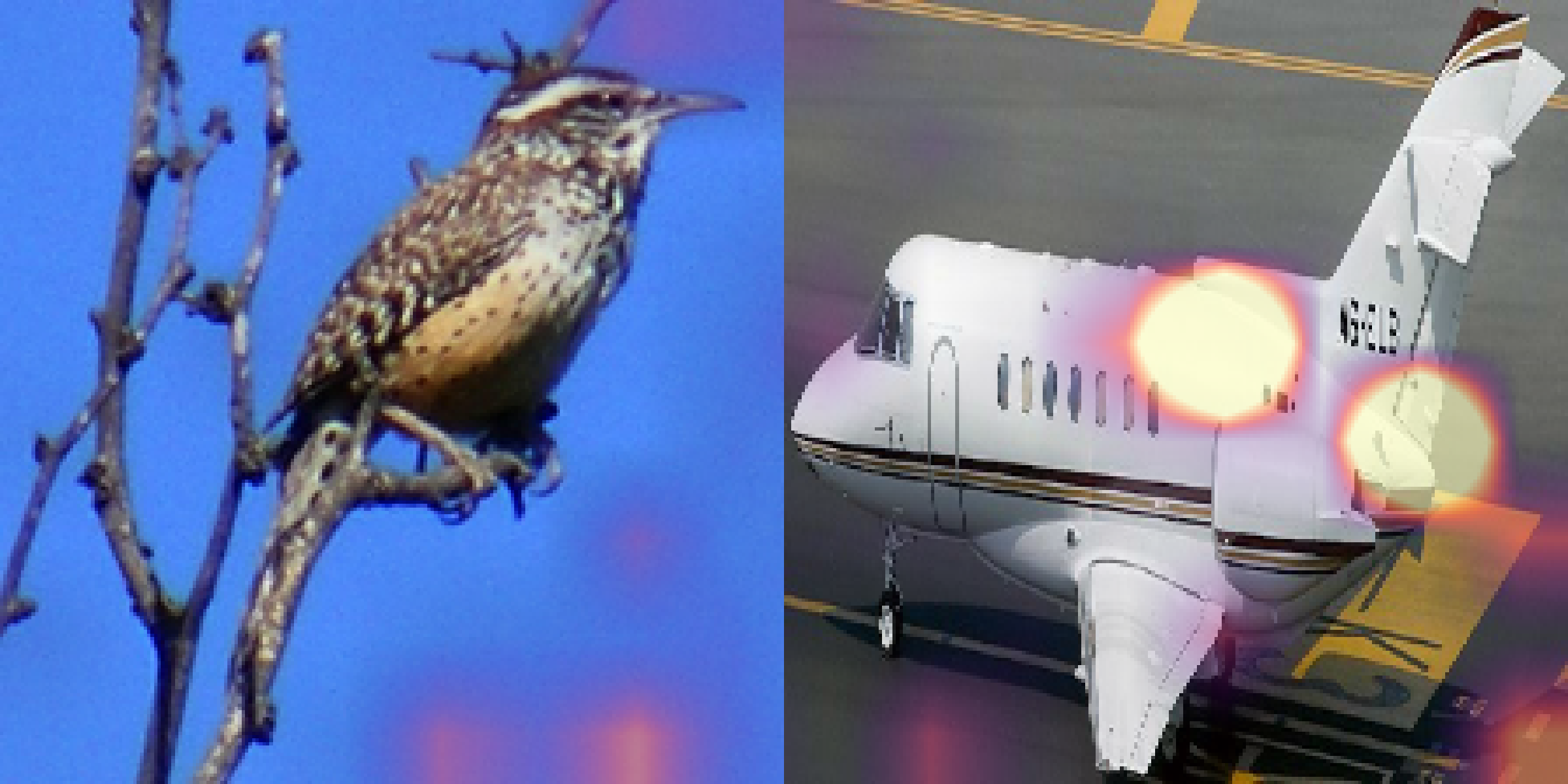} \\
    \vspace{-4pt} & \vspace{-4pt} \\
    \includegraphics[width=0.35\textwidth]{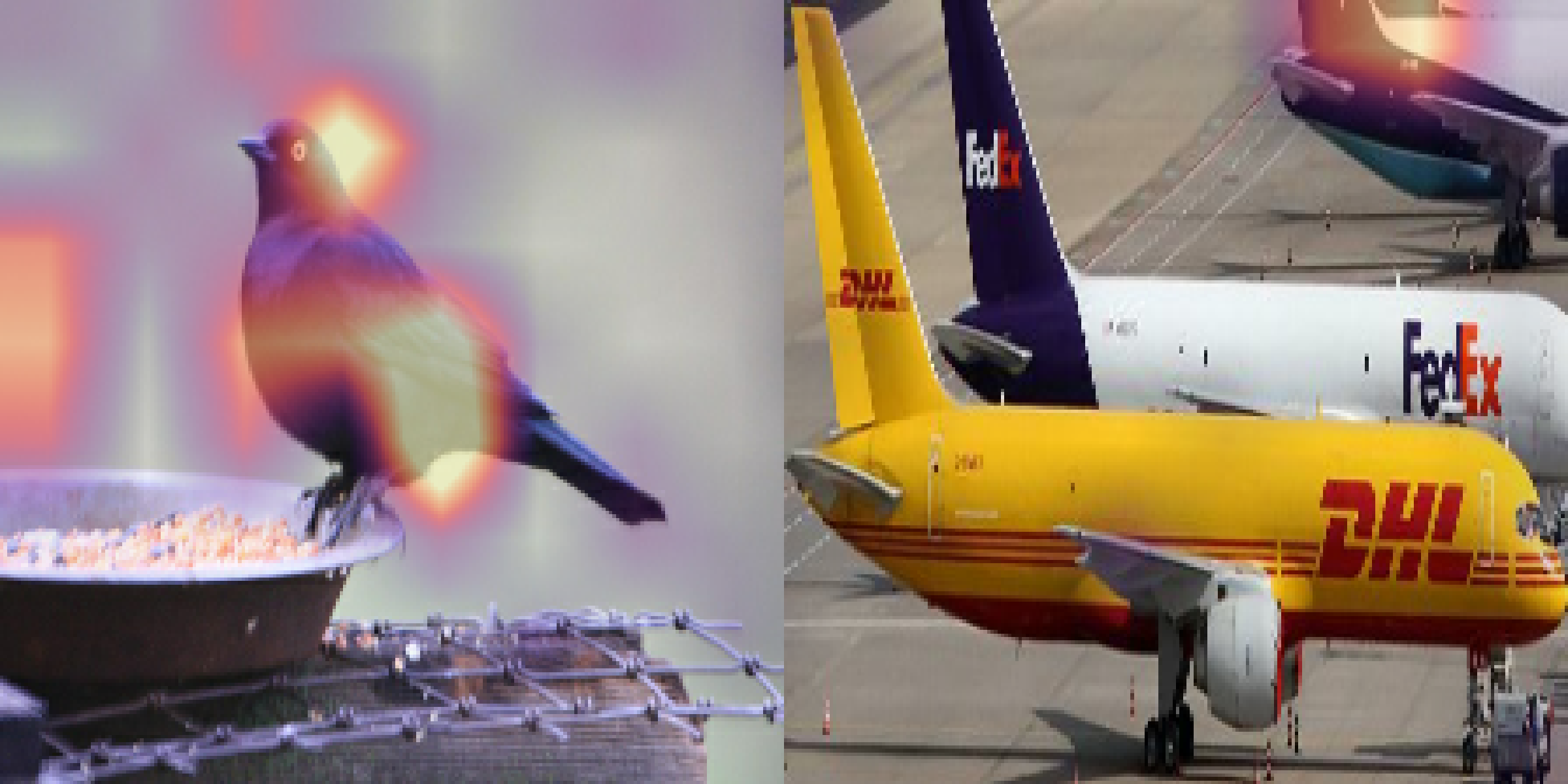} &
    \includegraphics[width=0.35\textwidth]{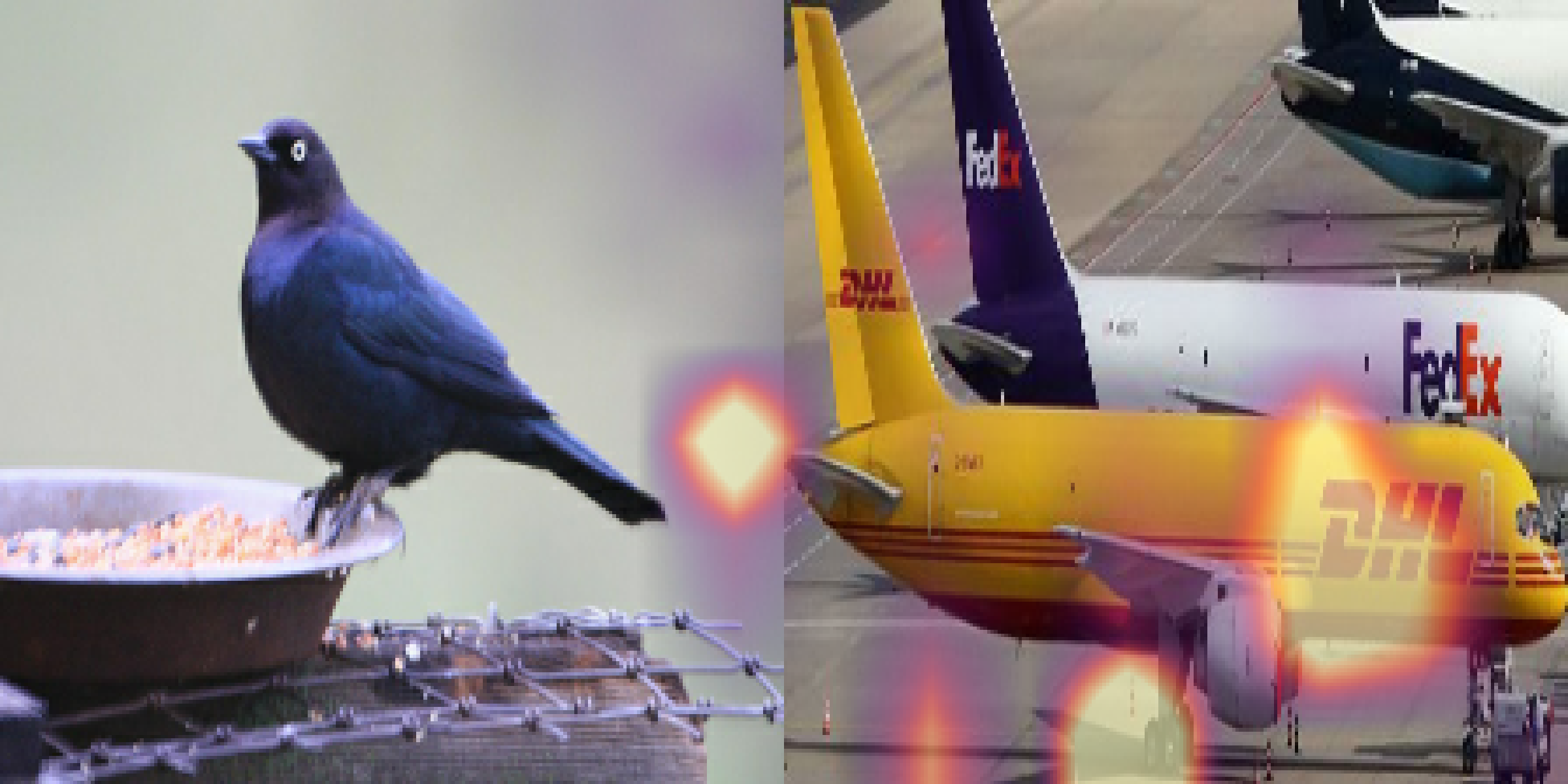} \\
    \vspace{-4pt} & \vspace{-4pt} \\
    {Prompt: \texttt{What type of bird is this?}} &
    {Prompt: \texttt{What type of plane is this?}}
  \end{tabular}
  \vspace{-4pt}
  \caption{
  \textbf{Top head attention map.}
  We concatenate bird~\cite{MD_BIRDS} and aircraft~\cite{CD_aircraft} datasets images horizontally in one support set. We then select the top vision-head for bird classification using the prompt \texttt{What type of bird is this?} and do the same for plane using the prompt \texttt{What type of plane is this?}. The attention map of the bird (left) and plane (right) top vision-head is overlaid on top of the image.}
  \label{fig:head-interpretability}
  \label{fig:interpretabiliy}
\end{figure}




To study the impact of prompt conditioning we conducted a series of experiments by randomly selecting 1000 10-way 4-shot tasks across 10 datasets. 
We processed the 40 images of the support set through Qwen2-VL using a prompt with \textcolor{attn_color}{Class} conditioning.
We then measure few-shot and zero-shot accuracy at different locations in the model.
Given either a token-embedding support set $\{z_i,y_i\}_{i=1}^{NK}$ extracted from an intermediate LLM layer, or an attention vector support set $\{\mathbf h^{(\text{v})}_{i,m},y_i\}_{i=1}^{NK}$, we fit a ridge linear classifier~\cite{bishop2006prml} to measure the few-shot accuracy of each distribution on the query set~\cite{alain2016understanding}.
Given class attention vectors $\{\mathbf h^{(\mathrm{t})}_{c,m}\}_{c\in\mathcal C}$, 
we measure the head zero-shot accuracy on the query set using as class logits $\mathbf h^{(\text{v})\top}_{q,m} \mathbf h^{(\mathrm{t})}_{c,m} $.
Results are shown in \cref{fig:qwen_probing_row}. We refer to the supplementary material for additional details on these experiments.

\Cref{fig:qwen_probing_a} shows the accuracy of the averaged vision tokens across the vision encoder and the LLM decoder.
As expected, as images are processed by the vision encoder, representations shift from low-level
to more discriminative high-level features, resulting in improved accuracy.
Once visual tokens are processed by the LLM decoder, their classification accuracy stalls.
Meanwhile, the last token representations refine from layer to layer by attending to the vision tokens and the prompt conditioning tokens, improving its accuracy.
As a result, the summary token yields higher accuracy than vision tokens, indicating that  LLM joint decoding of vision and prompt tokens steers representations during inference toward being more class-discriminative. \textbf{This demonstrates the LVLMs ability to refine visual features at inference using prompt conditioning.}


In \cref{fig:qwen_probing_c}, we select the 784 heads of the 28 layers of the last token. For each task, we rank them from best to worst accuracy and show their average few-shot accuracy.
We see that about 25\% of heads have better classification power than the LLM output, and that, more specifically, a handful of heads yields an accuracy gain of more than 10\% compared to the summary token. \textbf{This shows that for a given task, a sparse set of heads yields better performance at few-shot than the summary token itself.}

Similarly in \cref{fig:qwen_probing_d} we rank each head by its zero-shot classification accuracy and find that about 10 heads yield better zero-shot performance than the summary token. \textbf{This shows that for a given task, a sparse set of heads yields better performance at zero-shot than the summary token itself.}

\Cref{fig:head-interpretability} shows attention maps of the top vision-head for different \textcolor{domain_color}{Domain} prompts on a fixed support set. It shows that by combining prompt conditioning and top vision-head selection, we indeed retrieve domain-specific features.

\noindent In conclusion:
\begin{enumerate}[topsep=0pt]
    \item At inference time, the summary token gains in class separability thanks to prompt conditioning. This arises from the LLM decoder multimodal ability to jointly process vision tokens with text prompt tokens. 
    \item For each classification task, a sparse set of last token heads yield better performance than the summary token at zero-shot and few-shot.
\end{enumerate}
Therefore, identifying those top heads at test time has the potential to greatly increase few-shot and zero-shot performance of LVLMs.


\section{Method}

In this section, we propose a test-time ranking procedure to select the top vision and text heads. We then show how to combine predictions from those heads into a single classifier. An overview of the method is illustrated in \cref{fig:main}.



\subsection{Vision-Heads Ranking} 

Let us see how to rank the top vision-heads from a labeled support set.
Given the attention vector support set $\{\mathbf h^{(\text{v})}_{i,m},y_i\}_{i=1}^{NK}$ in a head $m$, the
classifier and its ranking score are both derived from a classical 
Gaussian Discriminant Analysis (GDA) \cite{bishop2006prml,CLIP_GDA}.
We assume a class-conditional generative model for the observed feature distribution.
Specifically, each attention vector follows a Gaussian distribution, centered at the class mean $\boldsymbol{\mu}_{m,c}\in\mathbb{R}^{D}$, with a covariance matrix $\boldsymbol{\Sigma}_m\in\mathbb{R}^{D\times D}$ that is shared across classes i.e.
\begin{equation}
    ({\mathbf H}_{m}^{(\text{v})}\mid Y=c)\sim\mathcal N\!\left(\boldsymbol\mu_{m,c},\boldsymbol\Sigma_m\right).
\end{equation}

A key reason this generative model performs well is that, in fine-grained few-shot classification, features often lie in a thin, anisotropic region of the embedding space.
The shared covariance among all classes captures principal directions common across classes, which improves discrimination.
In addition, with very few samples per class, class-specific covariance estimates are underconstrained, one shared covariance gives a more stable estimate from the full support set.


Given the observed features in a head, we estimate the mean $\hat{\boldsymbol\mu}_{m,c}$ and the precision matrix $\widehat{\boldsymbol\Sigma}_m^{-1}$ using the unbiased sample mean estimator and the empirical Bayes ridge-type estimator \cite{kubokawa2008estimation} respectively:
\begin{equation}
\hat{\boldsymbol\mu}_{m,c} = \frac{1}{K}\sum_{i:\,y_i=c}{\mathbf h}_{i,m}^{(\text{v})},
\qquad
\widehat{\boldsymbol\Sigma}_m^{-1}=D\Bigl((KN-1)\widehat{\boldsymbol\Sigma}_m+\mathrm{tr}(\widehat{\boldsymbol\Sigma}_m)\mathbf I_D\Bigr)^{-1},
\end{equation}
%
where $\mathbf I_D$ is the identity matrix and $\widehat{\boldsymbol\Sigma}_m$ is the empirical covariance. The class logits $\ell_{i,m,c}$ of the model are the log-probabilities of observing the features and the class label 
\begin{equation}
\begin{aligned}
\ell_{i,m,c}
&= \log p\!\left({\mathbf h}_{i,m}^{(\text{v})}, y=c\right)
= \log p\!\left({\mathbf h}_{i,m}^{(\text{v})}\mid y=c\right) + \log p(y=c) \\
&= -\frac{1}{2}\bigl({\mathbf h}^{(\text{v})}_{i,m}-\hat{\boldsymbol\mu}_{m,c}\bigr)^{\!\top}\widehat{\boldsymbol\Sigma}_m^{-1}\bigl({\mathbf h}_{i,m}^{(\text{v})}-\hat{\boldsymbol\mu}_{m,c}\bigr)
   + C,
\end{aligned}
\end{equation}
where the constant $C$ cancels out in the softmax. 
We compute class probabilities $p_{i,m,c}^{(\text{v})}$ via a temperature-scaled softmax, and define the vision-head score as 
\begin{equation}
s_m^{(\text{v})}
=
\frac{1}{KN}\sum_{i=1}^{KN} p_{i,m,y_i}^{(\text{v})}, 
\qquad\text{with}\qquad
p_{i,m,c}^{(\text{v})}
=
\frac{\exp\!\bigl(\ell_{i,m,c}/\tau\bigr)}{\sum_{j=1}^{N}\exp\!\bigl(\ell_{i,m,j}/\tau\bigr)}.
\end{equation}


This score can be interpreted as a soft accuracy on the support set, since in the limit $\tau \to 0$, $s_m^{(\text{v})}$ is the support set accuracy.
In the few-shot setting, the Gaussian model often overfits the support set.
Hence, the support set accuracy saturates at 100\% for many heads, 
reducing the ability to distinguish top heads from the others. In such cases, using softmax probabilities prevents $s_m^{(\mathrm{v})}$ from saturating.
This score is inspired by prior work on neural checkpoint ranking~\cite{li2021ranking,wang2023farranking_maha}, which aims to rank model transferability instead.
We define $\mathcal H^V$ as the set of top $k$ vision-heads according to the ranking score.

\subsection{Text-Heads Ranking}

Given the class attention vectors for the $m$-th text-head $\{\mathbf h^{(\mathrm{t})}_{c,m}\}_{c\in\mathcal C}$,
ranking the heads is straightforward as we simply evaluate the zero-shot soft accuracy of each head on the support set.
More specifically, for each support, 
we compute class logits as the dot product $\mathbf h_{i,m}^{(\text{v})\top}{\mathbf h}_{c,m}^{(\text{t})}$ and compute class probabilities by applying a softmax. The head score $s_{m}^{(\text{t})}$ is 
the average probability assigned to the ground-truth label over the support set:
\begin{equation}
s_{m}^{(\text{t})}
=
\frac{1}{KN}\sum_{i=1}^{KN} p_{i,m,y_i}^{(\text{t})},
\qquad
p_{i,m,c}^{(\text{t})}
=
\frac{\exp\!\left(\mathbf h_{i,m}^{(\text{v})\top}{\mathbf h}_{c,m}^{(\text{t})}\right)}
{\sum_{j=1}^{N}\exp\!\left(\mathbf h_{i,m}^{(\text{v})\top}{\mathbf h}_{j,m}^{(\text{t})}\right)}. 
\end{equation}
%
The set of top $k$ text-heads according to the ranking score $s_{m}^{(\text{t})}$ is $\mathcal H^T$.
It is important to note that this ranking needs labels, which are not available in pure zero-shot scenarios. 
However, in the vision-text-few-shot setup, a labeled set is available, making our method capable of training-free zero-shot adaptation. 
Moreover, experiments show that text-heads are shared across tasks and domains, hence a fixed $\mathcal H^T$, determined once per model, transfers across tasks.

\subsection{Head Ensemble Classifiers}

Given $\mathcal H^V$ and $\mathcal H^T$, we introduce 3 classifiers. 
HEC-V averages class probabilities of vision-heads to produce a vision-few-shot classifier and
HEC-T averages class probabilities of text-heads to produce a text-zero-shot classifier:
\begin{equation}
\bar{p}^{(\mathrm{HEC\text{-}V})}_{q,c}
=
\frac{1}{|\mathcal H^V|}\sum_{m\in\mathcal H^V} p_{q,m,c}^{(\text{v})},
\qquad
\bar{p}^{(\mathrm{HEC\text{-}T})}_{q,c}
=
\frac{1}{|\mathcal H^T|}\sum_{m\in\mathcal H^T} p_{q,m,c}^{(\text{t})}.
\end{equation}

Lastly, HEC-VT adds HEC-V and HEC-T class probabilities to produce a vision-text-few-shot classifier
\begin{equation}
\bar{p}^{(\mathrm{HEC\text{-}VT})}_{q,c}
=
\frac{\alpha\,\bar{p}^{(\mathrm{HEC\text{-}V})}_{q,c}
+\bar{p}^{(\mathrm{HEC\text{-}T})}_{q,c}}{\alpha + 1},
\end{equation}
where $\alpha$ is a hyper-parameter. Despite its simplicity, we find that ensembling heads by averaging class probabilities yields strong performance. 
It is robust to poorly ranked heads without adding extra hyperparameters or computations.

%% file: sec/3_exp.tex
\section{Experiments}

In this section we evaluate our methods on 12 datasets in three different setups.
We first benchmark HEC-V on \textbf{vision-few-shot} with \textcolor{domain_color}{Domain} conditioning, when class names are unknown.
We then benchmark HEC-T on \textbf{text-zero-shot} with \textcolor{attn_color}{Class} conditioning.
Lastly, we benchmark HEC-VT on \textbf{vision-text-few-shot} using \textcolor{domain_color}{Domain} conditioning.
We also include additional experiments on prompt conditioning and head selection.

\vspace{-.5em}\paragraph{Dataset.}
Following previous works~\cite{CLIP_CLIP,CLIP_GDA,CLIP_Proker},
we use 10 publicly available image classification datasets across different domains, covering a diverse range of visual recognition problems: 
EuroSAT (ESAT)~\cite{CD_eurosat}, 
UCF101 (UCF)~\cite{CD_ucf101}, 
DTD~\cite{CD_DTD}, 
Caltech101 (CAL)~\cite{CD_calthech_101},
SUN397 (SUN)~\cite{CD_sun},
OxfordPets (PETS)~\cite{CD_oxford_pets}, 
StanfordCars (CARS)~\cite{CD_stanford_cars}, 
Flowers102\! (FLWR)~\cite{CD_flowers_102}, 
Food101\! (FOOD)~\cite{CD_food_101}, and 
FGVCAircraft\! (FGVC)~\cite{CD_aircraft}. The last 5 are fine-grained image classification benchmarks.\\
We include in our experiments two additional fine-grained image classification datasets CUB-200 (BIRD)~\cite{MD_BIRDS} and Traffic-Signs (SIGN)~\cite{MD_Traffic}.
We treat ImageNet~\cite{CD_imagenet} as a general classification dataset, non domain-specific, and use it to select the top vision and text-heads shared across domains.

\vspace{-.5em}\paragraph{Protocol.}
We compare against state-of-the-art training-free CLIP-based baselines: 
closed-form linear probing~\cite{bishop2006prml} (Probing), CLIP~\cite{CLIP_CLIP}, TipAdapter~\cite{CLIP_tipadapter}, \\ GDA~\cite{CLIP_GDA}, and ProKeR~\cite{CLIP_Proker}.
Prior work reports results using the original OpenAI CLIP~\cite{CLIP_CLIP}.
Our method, paired with recent LVLMs, significantly outperforms these baselines, in part because recent LVLMs inherit recent and stronger CLIP backbones than OpenAI CLIP.
For a fair comparison, we try to disentangle the backbone performance from the contribution of our method. 
We therefore evaluate on two LVLMs where the pretrained CLIP backbone they inherit from is known. 
More specifically Qwen2-VL (7B)~\cite{qwen2VL} and LLaVA-OV (7B)~\cite{li2024llavaOV} use respectively DFN~\cite{qwen2VL_clip} and SigLIP~\cite{SigLIP} before instruction tuning. 
For LVLM-based training-free adaptation, we compare our work to state-of-the-art SAVs~\cite{SAV}.

All CLIP-based methods are evaluated using the same prompts originally introduced by TipAdapter~\cite{CLIP_tipadapter}. 
All LVLM-based methods are evaluated with the same prompts (see
details in the supplementary material).
All results are averaged over 5 random seeds.
For every dataset, we select the optimal set of hyperparameters 
using the original hyperparameter sweep used by each method. 
For linear probing, we use a ridge classifier and sweep the regularization coefficient with values ranging from 0.001 to 10.
To show the robustness of our method, we set $\tau=10$ and set to 20 the number of top vision-heads and the number of top text-heads we select across all models and benchmarks. 
Only for HEC-VT, we sweep $\alpha$ from 0.1 to 10.
All experiments are done on a single NVIDIA V100 GPU.


\input{table/sota_V}
\subsection{Vision-Few-Shot Classification}

We benchmark our method in the vision-few-shot setting, where class names are unknown.
We evaluate in $N$-way 4-shot with $N$ equal to the total number of classes in each dataset. The results are reported in \cref{tab:vision-few-shot-classification}.
For LVLM-based methods, we use \textcolor{domain_color}{Domain} prompt conditioning (\textcolor{domain_color}{DC}) e.g., \texttt{What breed is that
dog?}, except for linear probing where we additionally test \textcolor{task_color}{Task} prompt conditioning (\textcolor{task_color}{TC}) i.e., \texttt{What object is in the image?}.
This allows to extract more domain-specific features, improving class separability, which is impossible for vision models and CLIP models. 
Hence, we also evaluate several vision models and CLIP models using linear probing. 
We compare DFN~\cite{qwen2VL_clip}, OpenAI CLIP~\cite{CLIP_CLIP}, OpenCLIP~\cite{OpenCLIP} for CLIP models. 
We also compare against the DINO series of vision models DINOv1~\cite{caron2021dino}, DINOv2~\cite{oquab2023dinov2}, and DINOv3~\cite{simeoni2025dinov3}.

HEC-V achieves the best average accuracy 82.4\%, 
surpassing all LVLM-based methods including SAVs~\cite{SAV} across all datasets.
Compared to the strongest non-LVLM baselines HEC-V is best on 9/12 datasets, outperforming its vision backbone DFN on all datasets except EuroSAT. 
The strong results of HEC-V against DINOv3, despite using no hyperparameter tuning and relying on a less recent backbone, signals a promising new direction for training-free vision-few-shot classification. 
%
Qwen2-VL Probing(\textcolor{task_color}{TC}) is already competitive with strong visual backbones. However, Probing(\textcolor{domain_color}{DC}) adds 4\% in accuracy, confirming the hypothesis that domain conditioning helps retrieve domain-specific features.


\subsection{Text-Zero-Shot Classification}

We benchmark our method in the text-zero-shot setting. The results are reported in \cref{tab:text-zero-shot}.
For the baseline, we follow the standard LVLM zero-shot protocol~\cite{Finedefics}, framing classification as next-token prediction with a prompt that associates a letter with each class (e.g., \texttt{A: boxer, B: yorkshire terrier, C: golden retriever, ...}). 
As a stronger baseline, we also report the summary token (ST) zero-shot accuracy \eqref{eq:lvlm_encode_vision}\eqref{eq:lvlm_encode_text} following \cite{ouali2025vladva_cliplike}.
Lastly, we evaluate HEC-T in the zero-shot setup. 
HEC-T requires a labeled support set to select text-heads so we perform the head selection only once using the average ranking score over 100 randomly selected ImageNet tasks. We use that fixed set of 20 heads for all datasets. 
HEC-T and ST use a prompt with \textcolor{attn_color}{Class} conditioning.
For each dataset, we report the average over 100 10-way 0-shot tasks, to fit all classes into the prompt without degrading the performance of the baseline.
To verify the generality of the method, we test on two LVLMs: Qwen2-VL and LLaVA-OV.

\begin{table*}[t]
  \centering
  \caption{
  \textbf{Zero-Shot} classification accuracy (\%) on 10-way 0-shot across 12 datasets. 
  Our method HEC-T provides a training-free zero-shot adaptation that outperforms previous baselines and is competitive with CLIP backbones that LVLMs inherit from.
  HEC-T provides meaningful gain while enabling to zero-shot an unlimited number of classes.
  \underline{Underline} denotes the best LVLM method.
  \textbf{Bold} denotes the best including its CLIP backbone.
    \textcolor{darkgreen}{Green} denotes the absolute gain of HEC-T over the LVLM Baseline.}
    \vspace{-.5em}
    
\resizebox{\textwidth}{!}{%
  \setlength{\tabcolsep}{3pt}
  \begin{tabular}{ll|*{12}{>{\centering\arraybackslash}p{0.8cm}}|>{\centering\arraybackslash}p{0.8cm}}
    \toprule
    \shortstack[l]{Model} & \shortstack[l]{Method} & \shortstack[l]{PETS} & \shortstack[l]{ESAT} & \shortstack[l]{UCF} & \shortstack[l]{SUN} & \shortstack[l]{CAL} & \shortstack[l]{DTD} & \shortstack[l]{AIR} & \shortstack[l]{FOOD} & \shortstack[l]{FLWR} & \shortstack[l]{CARS} & \shortstack[l]{BIRD} & \shortstack[l]{SIGN} & \shortstack[l]{AVG} \\
    \midrule
    DFN & Zero-Shot & $\mathbf{97.6}$ & $53.0$ & $87.8$ & $\mathbf{97.4}$ & $\mathbf{99.3}$ & $78.4$ & $70.7$ & $\mathbf{96.7}$ & $\mathbf{93.8}$ & $\mathbf{99.6}$ & $\mathbf{96.5}$ & $45.3$ & $84.7$ \\
    \midrule
    \multirow{4}{*}{{\makecell{ DFN+LLM\\(Qwen2-VL)}}} %
      & Baseline & $84.1$ & $33.6$ & $85.6$ & $94.2$ & $98.4$ & $70.9$ & $62.1$ & $91.2$ & $78.3$ & $91.1$ & $69.7$ & $54.5$ & $76.1$ \\
     & ST~\cite{ouali2025vladva_cliplike} & $90.1$ & $41.6$ & $90.5$ & $94.5$ & $98.8$ & $77.3$ & $66.4$ & $92.3$ & $90.0$ & $96.0$ & $78.5$ & $56.8$ & $81.1$ \\
     & \hec{HEC-T} & 
     \hec{$\underline{95.2}$} & 
     \hec{$\underline{\mathbf{54.0}}$} & 
     \hec{$\underline{\mathbf{92.6}}$} & 
     \hec{$\underline{97.0}$} & 
     \hec{$\underline{99.3}$} & 
     \hec{$\underline{\mathbf{84.2}}$} & 
     \hec{$\underline{\mathbf{78.7}}$} & 
     \hec{$\underline{95.2}$} & 
     \hec{$\underline{91.1}$} & 
     \hec{$\underline{97.7}$} & 
     \hec{$\underline{85.9}$} & 
     \hec{$\underline{\mathbf{63.8}}$} & 
     \hec{$\underline{\mathbf{86.2}}$} \\
     &  & $\textcolor{darkgreen}{\raisebox{0.2ex}{\scriptsize +}11.1}$ & $\textcolor{darkgreen}{\raisebox{0.2ex}{\scriptsize +}20.4}$ & $\textcolor{darkgreen}{\raisebox{0.2ex}{\scriptsize +}7.0}$ & $\textcolor{darkgreen}{\raisebox{0.2ex}{\scriptsize +}2.8}$ & $\textcolor{darkgreen}{\raisebox{0.2ex}{\scriptsize +}0.9}$ & $\textcolor{darkgreen}{\raisebox{0.2ex}{\scriptsize +}13.3}$ & $\textcolor{darkgreen}{\raisebox{0.2ex}{\scriptsize +}16.6}$ & $\textcolor{darkgreen}{\raisebox{0.2ex}{\scriptsize +}4.0}$ & $\textcolor{darkgreen}{\raisebox{0.2ex}{\scriptsize +}12.9}$ & $\textcolor{darkgreen}{\raisebox{0.2ex}{\scriptsize +}6.7}$ & $\textcolor{darkgreen}{\raisebox{0.2ex}{\scriptsize +}16.2}$ & $\textcolor{darkgreen}{\raisebox{0.2ex}{\scriptsize +}9.3}$ & $\textcolor{darkgreen}{\raisebox{0.2ex}{\scriptsize +}10.1}$ \\
    \midrule
     
    \midrule
    SigLIP & Zero-Shot & $\mathbf{97.0}$ & $\mathbf{41.1}$ & $87.1$ & $96.8$ & $\mathbf{99.5}$ & $84.1$ & $\mathbf{79.6}$ & $\mathbf{97.1}$ & $\mathbf{95.8}$ & $\mathbf{99.5}$ & $\mathbf{95.5}$ & $47.1$ & $\mathbf{85.0}$ \\
    \midrule
    
    \multirow{4}{*}{{\makecell{SigLIP+LLM\\(LLaVA-OV)}}}
     & Baseline & $79.6$ & $37.2$ & $92.1$ & $96.0$ & $98.6$ & $77.8$ & $61.9$ & $94.8$ & $66.7$ & $92.5$ & $63.7$ & $72.6$ & $79.7$ \\
    & ST~\cite{ouali2025vladva_cliplike} & $69.8$ & $32.1$ & $91.7$ & $52.4$ & $95.9$ & $44.2$ & $64.7$ & $69.3$ & $50.0$ & $93.6$ & $59.1$ & $59.5$ & $65.2$ \\
    & \hec{HEC-T} & 
    \hec{$\underline{83.8}$} & 
    \hec{$\underline{40.0}$} & 
    \hec{$\underline{\mathbf{94.2}}$} & 
    \hec{$\underline{\mathbf{97.2}}$} & 
    \hec{$\underline{99.1}$} & 
    \hec{$\underline{\mathbf{84.5}}$} & 
    \hec{$\underline{73.0}$} & 
    \hec{$\underline{95.4}$} & 
    \hec{$\underline{72.2}$} & 
    \hec{$\underline{96.8}$} & 
    \hec{$\underline{67.2}$} & 
    \hec{$\underline{\mathbf{73.7}}$} & 
    \hec{$\underline{82.1}$} \\
    & \textcolor{darkgreen}{} & $\textcolor{darkgreen}{\raisebox{0.2ex}{\scriptsize +}4.2}$ & $\textcolor{darkgreen}{\raisebox{0.2ex}{\scriptsize +}2.8}$ & $\textcolor{darkgreen}{\raisebox{0.2ex}{\scriptsize +}2.1}$ & $\textcolor{darkgreen}{\raisebox{0.2ex}{\scriptsize +}1.3}$ & $\textcolor{darkgreen}{\raisebox{0.2ex}{\scriptsize +}0.5}$ & $\textcolor{darkgreen}{\raisebox{0.2ex}{\scriptsize +}6.7}$ & $\textcolor{darkgreen}{\raisebox{0.2ex}{\scriptsize +}11.1}$ & $\textcolor{darkgreen}{\raisebox{0.2ex}{\scriptsize +}0.6}$ & $\textcolor{darkgreen}{\raisebox{0.2ex}{\scriptsize +}5.5}$ & $\textcolor{darkgreen}{\raisebox{0.2ex}{\scriptsize +}4.3}$ & $\textcolor{darkgreen}{\raisebox{0.2ex}{\scriptsize +}3.5}$ & $\textcolor{darkgreen}{\raisebox{0.2ex}{\scriptsize +}1.1}$ & $\textcolor{darkgreen}{\raisebox{0.2ex}{\scriptsize +}2.4}$ \\
    \bottomrule
  \end{tabular}
\label{tab:text-zero-shot}
}
\end{table*}

\input{table/sota_VT}

On average, HEC-T improves Qwen2-VL zero-shot by +10.1\% surpassing its backbone (DFN) by 1.5\% on average.
For both Qwen2-VL and LLaVA-OV, HEC-T outperforms ST and the baseline on every dataset.
For LLaVA-OV, HEC-T yields a smaller gain of +2.4\% over the baseline, but improves ST by +16.9\%.
Notably, ST and HEC-T are the only methods that can scale with the number of classes as the baseline is limited by the context window.
HEC-T's consistent gains across 12 heterogeneous benchmarks support that top text-heads transfer across domains.
However, the CLIP backbones still win on more datasets overall (DFN beats Qwen2-VL HEC-T on 7/12 datasets; SigLIP beats LLaVA-OV HEC-T on 8/12). While HEC-T bridged the gap between LVLMs and CLIPs in zero-shot scenarios, CLIP still yielded strong performance.
We notice that in general, CLIP wins on saturated benchmarks. For Qwen2-VL, HEC-T wins only when the performance is below 90\%. 
This hints that LVLMs with HEC-T are more robust to domains under-represented in pretraining data.


\subsection{Vision-Text-Few-Shot Classification}

We benchmark our method in the vision-text-few-shot setting.
We evaluate in $N$-way 4-shot with $N$ equal to the total number of classes in the dataset. The results are reported in \cref{tab:fewshot_VT}.
We evaluate all baselines using CLIP and LVLM as an encoder \eqref{eq:lvlm_encode_vision} \eqref{eq:lvlm_encode_text}.
For LVLM-based methods, we use \textcolor{domain_color}{Domain} prompt conditioning. 
We do not include classes in the prompt, as most datasets have $N \gg 20$ classes, which would cause a drop in performance.

HEC-VT outperforms all LVLM-based baselines by more than 3\% on average. 
Combining HEC-T and HEC-V improves performance on every dataset except UCF that already yields strong results with HEC-V. 
Averaged over all datasets, HEC is the only LVLM method that surpasses the best CLIP-based baseline.
However, CLIP-based methods still achieve higher accuracy on 5 out of the 12 datasets.
We hypothesize that part of that performance gap could be linked to the post-training of Qwen2-VL. 
Similarly to zero-shot, we notice that HEC-VT wins on less saturated benchmarks.
Additionally, HEC-VT consistently outperforms CLIP-based methods on less object-centric datasets, such as textures (DTD), scenes (SUN), and human actions (UCF).

\begin{figure*}[t]
    \centering

    \begin{minipage}[t]{0.5\linewidth}
        \vspace{0pt}
        \centering
        \begin{subtable}[t]{\linewidth}
            \centering
            \small
            \resizebox{\linewidth}{!}{%
            \begin{tabular}{l | c c | c c}
            \toprule
            \multicolumn{1}{c}{} & \multicolumn{2}{c}{HEC-T} & \multicolumn{2}{c}{HEC-V} \\
            Conditioning & Acc. & ER & Acc. & ER \\
            \midrule
            None & $82.34_{0.8}$ & - & $90.43_{0.5}$ & - \\
            \textcolor{task_color}{Task} & $84.09_{0.7}$  & $\textcolor{darkgreen}{\downarrow\,9.89\%}$ & $91.42_{0.5}$ & $\textcolor{darkgreen}{\downarrow\,10.31\%}$ \\
            \textcolor{domain_color}{Domain} & $87.08_{0.6}$ & $\textcolor{darkgreen}{\downarrow\,18.81\%}$ & $92.78_{0.4}$ & $\textcolor{darkgreen}{\downarrow\,15.85\%}$ \\
            \textcolor{attn_color}{Class} & $88.45_{0.6}$ & $\textcolor{darkgreen}{\downarrow\,10.63\%}$ & $94.14_{0.4}$ & $\textcolor{darkgreen}{\downarrow\,18.84\%}$ \\
            \bottomrule
            \end{tabular}
            }%
            \caption{
            10-way 4-shot performance under different conditioning.
            ER stands for Error Reduction in percentage.
            }
            \label{tab:task_domain_class}
        \end{subtable}
    \end{minipage}\hfill
    \begin{minipage}[t]{0.49\linewidth}
        \vspace{0pt}
        \centering

        \begin{subfigure}[t]{0.41\linewidth}
            \centering
            \includegraphics[height=2.5cm,keepaspectratio]{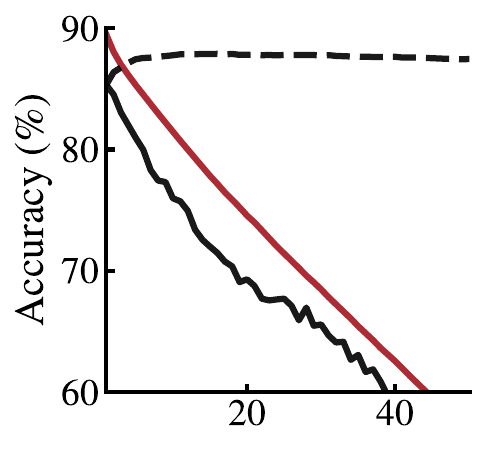}
            \caption{
            text-head zero-shot accuracy
            }
            \label{fig:heads_HEC-T}
        \end{subfigure}\hspace{1em}
        \begin{subfigure}[t]{0.41\linewidth}
            \centering
            \includegraphics[height=2.5cm,keepaspectratio]{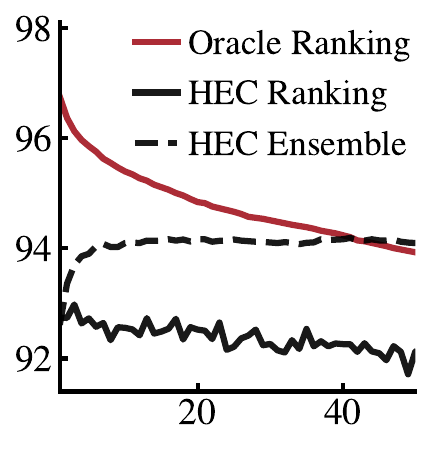}
            \caption{
            vision-head few-shot accuracy
            }
            \label{fig:heads_HEC-V}
        \end{subfigure}
    \end{minipage}
\vspace{-.5em}

    \caption{
    \textbf{Ablation studies.} Prompt Conditioning (left) and Head Ranking (right).
    }
    \label{fig:ablation_table_and_heads}
\end{figure*}

\subsection{Ablation Studies: Prompt Conditioning and Head Ranking}
\Cref{tab:task_domain_class} reports performance given four types of prompt conditioning: None, \textcolor{task_color}{Task}, \textcolor{domain_color}{Domain}, and \textcolor{attn_color}{Class}. Incrementally adding conditioning results in better zero-shot and few-shot performance.
Given the setup of \cref{sec:preliminaries}, \cref{fig:heads_HEC-T,fig:heads_HEC-V} report the performance of our ranking and ensemble method on the top 50 heads, showing HEC robustness. Details are reported in the supplementary material.


\section{Conclusion}
We've seen that HEC improves few-shot classification across a variety of setups, notably showcasing prompt-guided domain adaptation. 
In addition, it closes the performance gap between LVLM-based and CLIP-based methods without the need for fine-tuning.
Thus we think that combining prompt conditioning with top head selection has the potential to generalize to other setups beyond few-shot and zero-shot classification. Future work will focus on adding more complex prompts and in-context examples.
Our implementation and evaluation code will be publicly released.

\vspace{-.5em}\paragraph{Limitations}
We acknowledge that the need for an intermediate representation (i.e., $\mathbf{h}_m$) for classification is a limitation, especially for API-based usage.
Also, class conditioning, while promising, is limited to a small number of classes. Furthermore, LVLM inference is more computationally intensive than CLIP models.

%% file: table/sota_V.tex
\begin{table}[t]
  \centering
  \caption{
\textbf{Vision-Few-Shot} classification accuracy (\%) on 4-shot across 12 datasets without knowing class names. %
\textcolor{domain_color}{DC} indicates \textcolor{domain_color}{Domain} prompt conditioning and \textcolor{task_color}{TC} indicates \textcolor{task_color}{Task} prompt conditioning. %
HEC-V achieves state-of-the-art performance on average and across all datasets except FLWR, BIRD and ESAT. %
\underline{Underline} denotes the best LVLM method.
\textbf{Bold} denotes the best overall.
Methods marked with $^\dagger$ do not use hyperparameter tuning.
  }\label{tab:vision-few-shot-classification}
  \vspace{-.5em}
  
\resizebox{\textwidth}{!}{%
  \setlength{\tabcolsep}{3pt}
  \begin{tabular}{ll|*{12}{>{\centering\arraybackslash}p{0.8cm}}|>{\centering\arraybackslash}p{0.8cm}}
    \toprule
    \shortstack[l]{Model} & \shortstack[l]{Method} & \shortstack[l]{PETS} & \shortstack[l]{ESAT} & \shortstack[l]{UCF} & \shortstack[l]{SUN} & \shortstack[l]{CAL} & \shortstack[l]{DTD} & \shortstack[l]{AIR} & \shortstack[l]{FOOD} & \shortstack[l]{FLWR} & \shortstack[l]{CARS} & \shortstack[l]{BIRD} & \shortstack[l]{SIGN} & \shortstack[l]{AVG} \\
    \midrule
    DINOv1 & Probing & $81.9$ & $\mathbf{81.6}$ & $71.8$ & $50.8$ & $86.9$ & $49.6$ & $25.5$ & $37.9$ & $88.1$ & $28.3$ & $54.2$ & $46.3$ & $58.6$ \\
    DINOv2 & Probing & $78.5$ & $73.6$ & $67.6$ & $63.4$ & $87.0$ & $51.6$ & $29.9$ & $43.0$ & $97.0$ & $35.9$ & $65.5$ & $35.6$ & $60.7$ \\
    DINOv3 & Probing & $88.0$ & $77.2$ & $82.8$ & $72.7$ & $95.4$ & $63.6$ & $56.9$ & $74.3$ & $\mathbf{99.5}$ & $79.4$ & $\mathbf{77.3}$ & $53.1$ & $76.7$ \\
    OpenAI CLIP & Probing & $72.9$ & $73.6$ & $81.5$ & $73.2$ & $90.2$ & $55.8$ & $28.2$ & $73.4$ & $89.5$ & $56.9$ & $53.6$ & $56.5$ & $67.1$ \\
    OpenCLIP & Probing & $73.0$ & $75.6$ & $79.5$ & $72.1$ & $90.4$ & $60.3$ & $29.6$ & $62.9$ & $89.0$ & $75.0$ & $51.2$ & $62.7$ & $68.4$ \\
    DFN & Probing & $84.5$ & $80.8$ & $79.8$ & $73.9$ & $94.4$ & $62.8$ & $40.0$ & $77.5$ & $96.8$ & $85.6$ & $66.9$ & $68.2$ & $75.9$ \\
    \midrule
    
    \multirow{4}{*}{{\makecell{ DFN+LLM\\(Qwen2-VL)}}} %
    & Probing(\textcolor{task_color}{TC}) & $84.9$ & $75.6$ & $81.4$ & $77.2$ & $93.4$ & $58.2$ & $40.9$ & $81.1$ & $98.1$ & $79.8$ & $65.1$ & $71.3$ & $75.6$ \\
    & Probing(\textcolor{domain_color}{DC}) & $92.0$ & $74.0$ & $82.3$ & $79.8$ & $94.2$ & $66.9$ & $60.7$ & $82.7$ & $98.2$ & $89.2$ & $69.5$ & $69.2$ & $79.9$ \\
    & SAVs$^\dagger$(\textcolor{domain_color}{DC})~\cite{SAV} & $91.0$ & $72.0$ & $80.5$ & $81.7$ & $94.4$ & $70.5$ & $59.3$ & $84.9$ & $97.8$ & $89.5$ & $69.8$ & $68.1$ & $80.0$ \\
    & 
    \hec{HEC-V$^\dagger$(\textcolor{domain_color}{DC})} & 
    \hec{$\underline{\mathbf{92.2}}$} & 
    \hec{$\underline{78.8}$} & 
    \hec{$\underline{\mathbf{85.0}}$} & 
    \hec{$\underline{\mathbf{82.4}}$} & 
    \hec{$\underline{\mathbf{95.5}}$} & 
    \hec{$\underline{\mathbf{71.8}}$} & 
    \hec{$\underline{\mathbf{62.2}}$} & 
    \hec{$\underline{\mathbf{85.3}}$} & 
    \hec{$\underline{98.5}$} & 
    \hec{$\underline{\mathbf{89.8}}$} & 
    \hec{$\underline{72.0}$} & 
    \hec{$\underline{\mathbf{75.5}}$} & 
    \hec{$\underline{\mathbf{82.4}}$} \\
    \bottomrule
  \end{tabular}
}
\end{table}

%% file: table/sota_VT.tex
\begin{table*}[t]
\centering
\caption{
    \textbf{Vision-Text-Few-shot}  classification accuracy (\%) on 4-shot across 12 datasets. We report results for CLIP-based and LVLM-based baselines.
    Combining HEC-T and HEC-V in a single classifier gives state-of-the-art performance, outperforming previous CLIP-based or LVLM-based baselines.
    \underline{Underline} denotes the best LVLM-based method.
    \textbf{Bold} denotes the best overall.
    Methods marked with $^\dagger$ do not use hyperparameter tuning.
    }
\label{tab:fewshot_VT}
\vspace{-.5em}

\resizebox{\textwidth}{!}{%
  \setlength{\tabcolsep}{3pt}
  \begin{tabular}{ll|*{12}{>{\centering\arraybackslash}p{0.8cm}}|>{\centering\arraybackslash}p{0.8cm}}
  \toprule
    \shortstack[l]{Model} & \shortstack[l]{Method} & \shortstack[l]{PETS} & \shortstack[l]{ESAT} & \shortstack[l]{UCF} & \shortstack[l]{SUN} & \shortstack[l]{CAL} & \shortstack[l]{DTD} & \shortstack[l]{AIR} & \shortstack[l]{FOOD} & \shortstack[l]{FLWR} & \shortstack[l]{CARS} & \shortstack[l]{BIRD} & \shortstack[l]{SIGN} & \shortstack[l]{AVG} \\
    \midrule
    \multirow{5}{*}{\rotatebox[origin=c]{90}{\makecell{DFN}}}
     & Zero-Shot$^\dagger$~\cite{CLIP_CLIP} & $92.0$ & $51.6$ & $63.4$ & $79.5$ & $95.6$ & $51.1$ & $29.6$ & $87.2$ & $82.0$ & $92.1$ & $78.0$ & $28.1$ & $69.2$ \\
     & Probing~\cite{bishop2006prml}  & $84.9$ & $81.6$ & $79.9$ & $73.7$ & $94.4$ & $61.4$ & $38.0$ & $77.0$ & $97.3$ & $85.2$ & $66.7$ & $65.9$ & $75.5$ \\
     & TipAdapter~\cite{CLIP_tipadapter} & $92.3$ & $69.2$ & $77.4$ & $80.5$ & $95.8$ & $64.4$ & $40.2$ & $87.2$ & $97.0$ & $92.7$ & $78.7$ & $50.0$ & $77.1$ \\
     & GDA~\cite{CLIP_GDA} & $\mathbf{92.8}$ & $78.0$ & $84.0$ & $83.2$ & $96.5$ & $70.0$ & $46.3$ & $87.2$ & $98.5$ & $\mathbf{93.6}$ & $80.3$ & $67.8$ & $81.5$ \\
     & ProKeR~\cite{CLIP_Proker} & $91.2$ & $\mathbf{82.4}$ & $83.9$ & $82.2$ & $\mathbf{97.0}$ & $66.2$ & $43.1$ & $\mathbf{87.8}$ & $98.3$ & $\mathbf{93.6}$ & $\mathbf{81.5}$ & $64.6$ & $81.0$ \\

    \midrule
    \multirow{9}{*}{\rotatebox[origin=c]{90}{\makecell{DFN+LLM \\ (Qwen2-VL)}}}
     & Zero-Shot$^\dagger$~\cite{ouali2025vladva_cliplike} & $55.0$ & $48.8$ & $30.8$ & $65.5$ & $73.3$ & $32.0$ & $30.1$ & $72.4$ & $8.2$ & $45.5$ & $6.1$ & $30.9$ & $41.5$ \\
     & Probing~\cite{bishop2006prml} & $92.0$ & $74.0$ & $82.3$ & $79.8$ & $94.2$ & $66.9$ & $60.7$ & $82.7$ & $98.2$ & $89.2$ & $69.5$ & $69.2$ & $79.9$ \\
     & TipAdapter~\cite{CLIP_tipadapter} & $79.0$ & $50.4$ & $57.6$ & $76.1$ & $84.9$ & $58.5$ & $51.8$ & $79.0$ & $74.7$ & $74.6$ & $54.0$ & $49.0$ & $65.8$ \\
     & GDA~\cite{CLIP_GDA}& $92.3$ & $69.2$ & $79.5$ & $82.1$ & $94.0$ & $68.7$ & $60.4$ & $83.9$ & $96.4$ & $87.9$ & $69.4$ & $64.5$ & $79.0$ \\
     & ProKeR~\cite{CLIP_Proker} & $86.7$ & $74.4$ & $77.4$ & $81.4$ & $94.1$ & $67.2$ & $55.8$ & $84.4$ & $94.4$ & $86.9$ & $65.7$ & $63.3$ & $77.6$ \\
     & SAVs$^\dagger$~\cite{SAV} & $91.0$ & $72.0$ & $80.5$ & $81.7$ & $94.4$ & $70.5$ & $59.3$ & $84.9$ & $97.8$ & $89.5$ & $69.8$ & $68.1$ & $80.0$ \\
    \cmidrule(lr){2-15}
     & \hec{HEC-T}$^\dagger$ & 
     \hec{$85.2$} & 
     \hec{$55.6$} & 
     \hec{$69.3$} & 
     \hec{$75.1$} & 
     \hec{$92.4$} & 
     \hec{$62.6$} & 
     \hec{$31.6$} & 
     \hec{$83.9$} & 
     \hec{$50.7$} & 
     \hec{$72.6$} & 
     \hec{$47.1$} & 
     \hec{$47.9$} & 
     \hec{$64.5$} \\
     & \hec{HEC-V}$^\dagger$ & 
     \hec{$92.2$} & 
     \hec{$78.8$} & 
     \hec{$\underline{\mathbf{85.0}}$} & 
     \hec{$82.4$} & 
     \hec{$95.5$} & 
     \hec{$71.8$} & 
     \hec{$62.2$} & 
     \hec{$85.3$} & 
     \hec{$98.5$} & 
     \hec{$89.8$} & 
     \hec{$72.0$} & 
     \hec{$75.5$} & 
     \hec{$82.4$} \\
     & \hec{HEC-VT} & 
     \hec{$\underline{\mathbf{92.8}}$} & 
     \hec{$\underline{82.0}$} & 
     \hec{$\underline{\mathbf{85.0}}$} & 
     \hec{$\underline{\mathbf{83.3}}$} & 
     \hec{$\underline{95.6}$} & 
     \hec{$\underline{\mathbf{72.7}}$} & 
     \hec{$\underline{\mathbf{62.3}}$} & 
     \hec{$\underline{85.7}$} &
     \hec{$\underline{\mathbf{98.6}}$} & 
     \hec{$\underline{90.1}$} & 
     \hec{$\underline{72.1}$} & 
     \hec{$\underline{\mathbf{76.2}}$} & 
     \hec{$\underline{\mathbf{83.0}}$} \\
    \bottomrule

  \end{tabular}
  }
\end{table*}

%% file: sec/4_supp.tex
\appendix
\section*{Supplementary Material}

This supplementary material provides additional experimental details and analyses for the results presented in the main paper.

\Cref{sec:supp_IMPLEM} provides additional details of the experimental setup.
\begin{itemize}
    \item \Cref{sec:supp_prelim} provides the implementation details of the preliminary experiments.
    \item \Cref{sec:supp_eval} provides the implementation details of the evaluation protocol used throughout the experiments.
    \item \Cref{sec:supp_prompts} provides the prompts used in the experiments.
    \item \Cref{sec:supp_cost} analyzes the computational cost of HEC-V compared with linear probing.
\end{itemize}

\Cref{sec:supp_ABLATION} studies the main design choices of the method.
\begin{itemize}
    \item \Cref{sec:supp_ensemble} studies the ensemble method.
    \item \Cref{sec:supp_tau} studies the effect of the temperature hyperparameter $\tau$.
    \item \Cref{sec:supp_heads} studies the head selection mechanism.
    \item \Cref{sec:supp_failling} studies a failing case of Class conditioning.
\end{itemize}

\Cref{sec:supp_RESULT} reports complementary experimental results beyond the main setting.
\begin{itemize}
    \item \Cref{sec:supp_itc} reports additional experiments on image-text retrieval.
    \item \Cref{sec:supp_other_models} reports performance gain from HEC-VT using 3 other models.
\end{itemize}

Unless otherwise specified, all experiments are conducted using Qwen2-VL-7B~\cite{qwen2VL} with \textcolor{attn_color}{Class} conditioning on 10-way 4-shot tasks with $\tau=10$ and top-$k= 20$.
All reported uncertainties, written as subscripts such as \(_{0.6}\), denote 95\% uncertainty intervals.
\section{Implementation Details}\label{sec:supp_IMPLEM}

\subsection{Preliminaries}\label{sec:supp_prelim}
In this section, we provide more details on how the preliminary experiments were conducted.

\begin{figure}[t]
  \centering
  \setlength{\tabcolsep}{6pt} 
  \renewcommand{\arraystretch}{1.0}
  \begin{tabular}{@{}c c@{}}
    \includegraphics[width=0.35\textwidth]{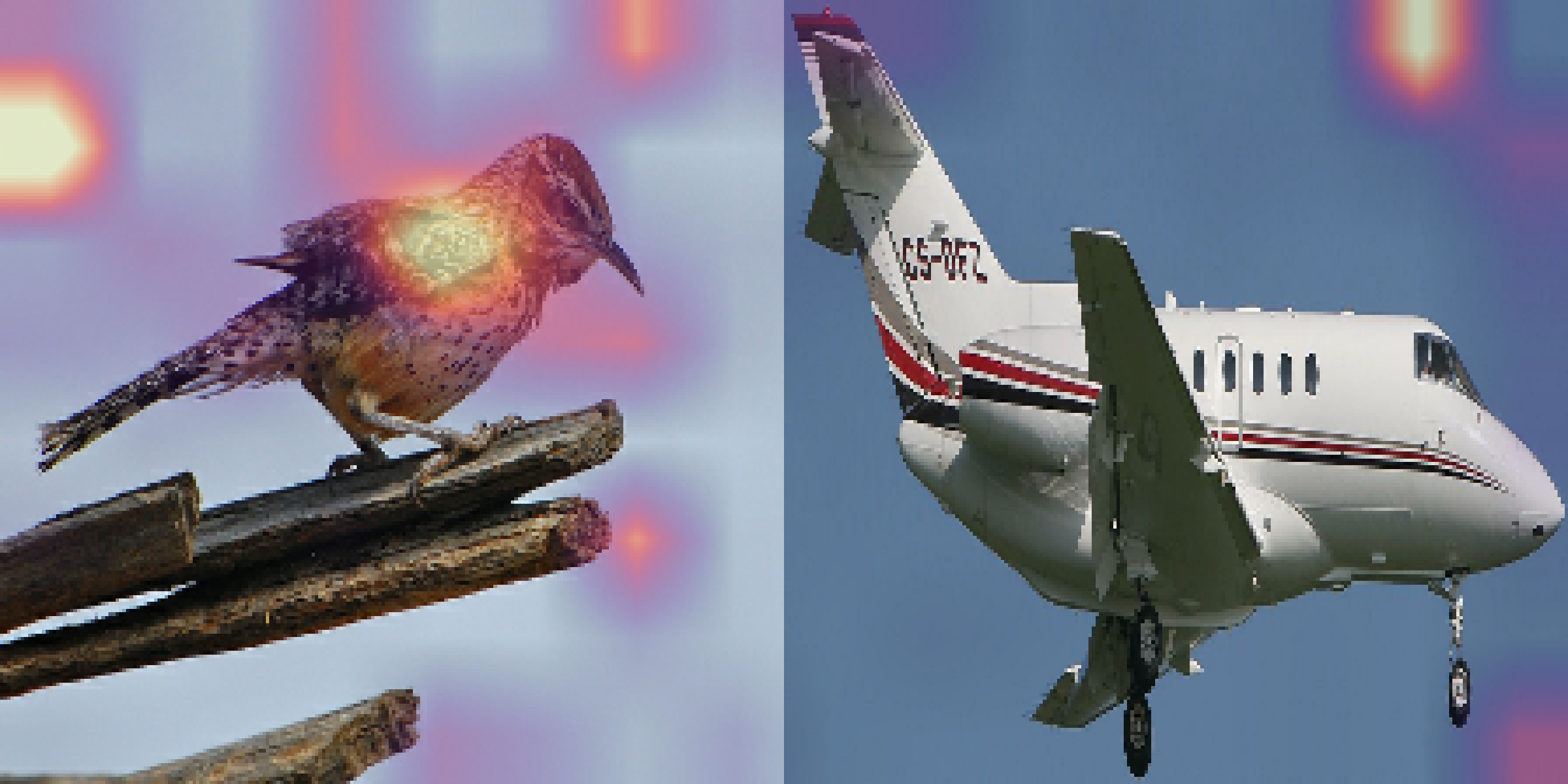} &
    \includegraphics[width=0.35\textwidth]{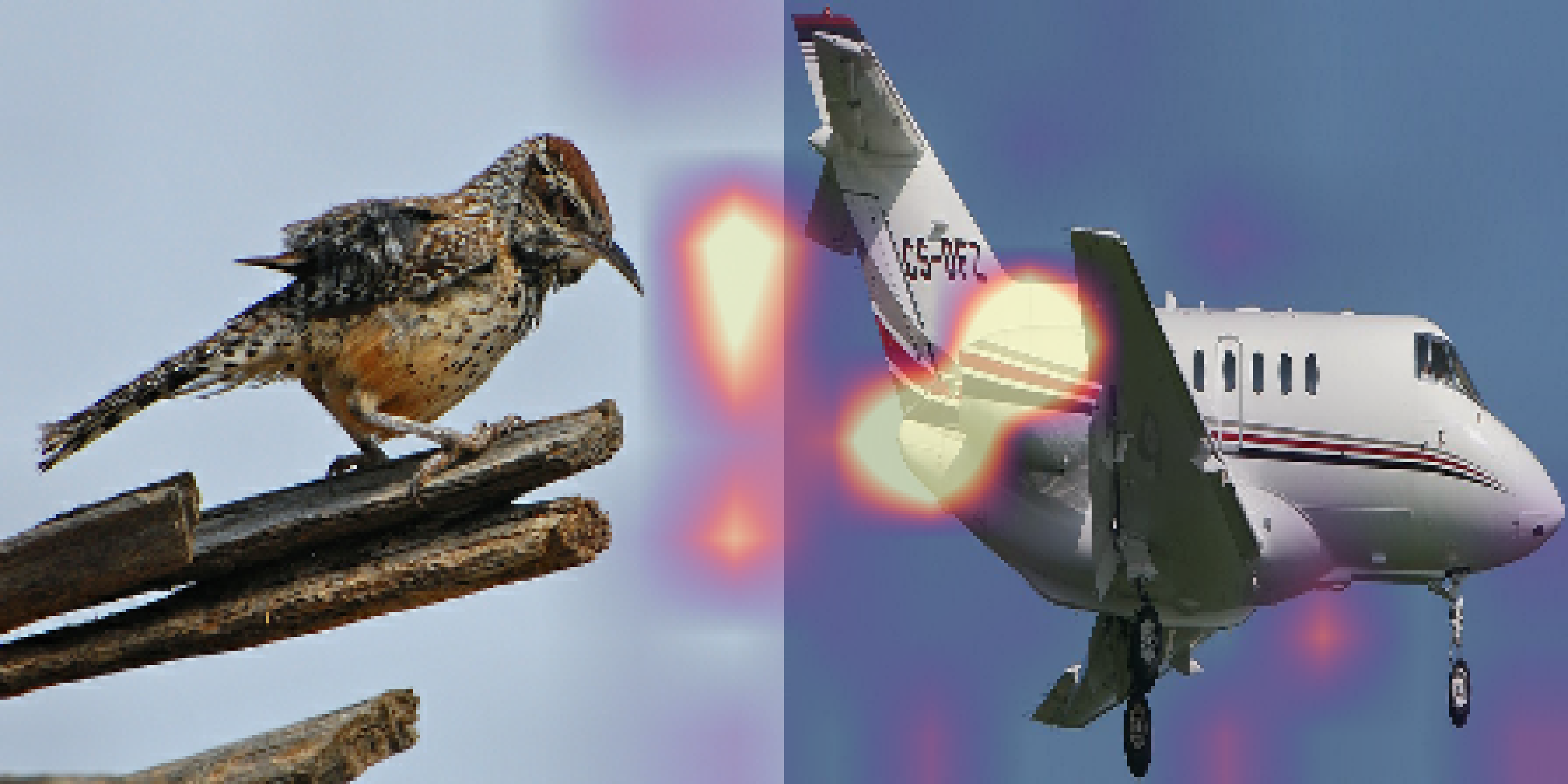} \\
    \vspace{-4pt} & \vspace{-4pt} \\
    \includegraphics[width=0.35\textwidth]{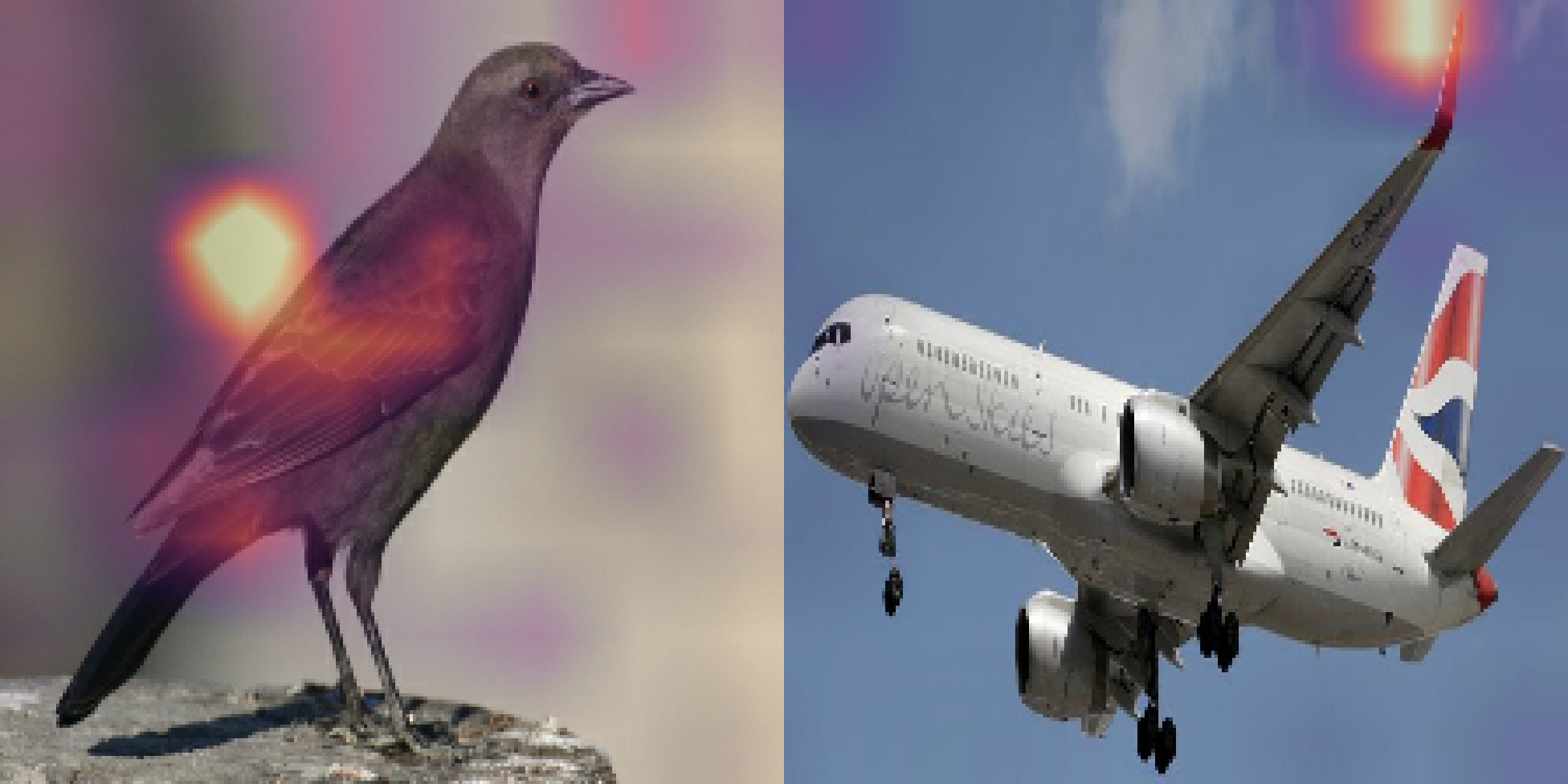} &
    \includegraphics[width=0.35\textwidth]{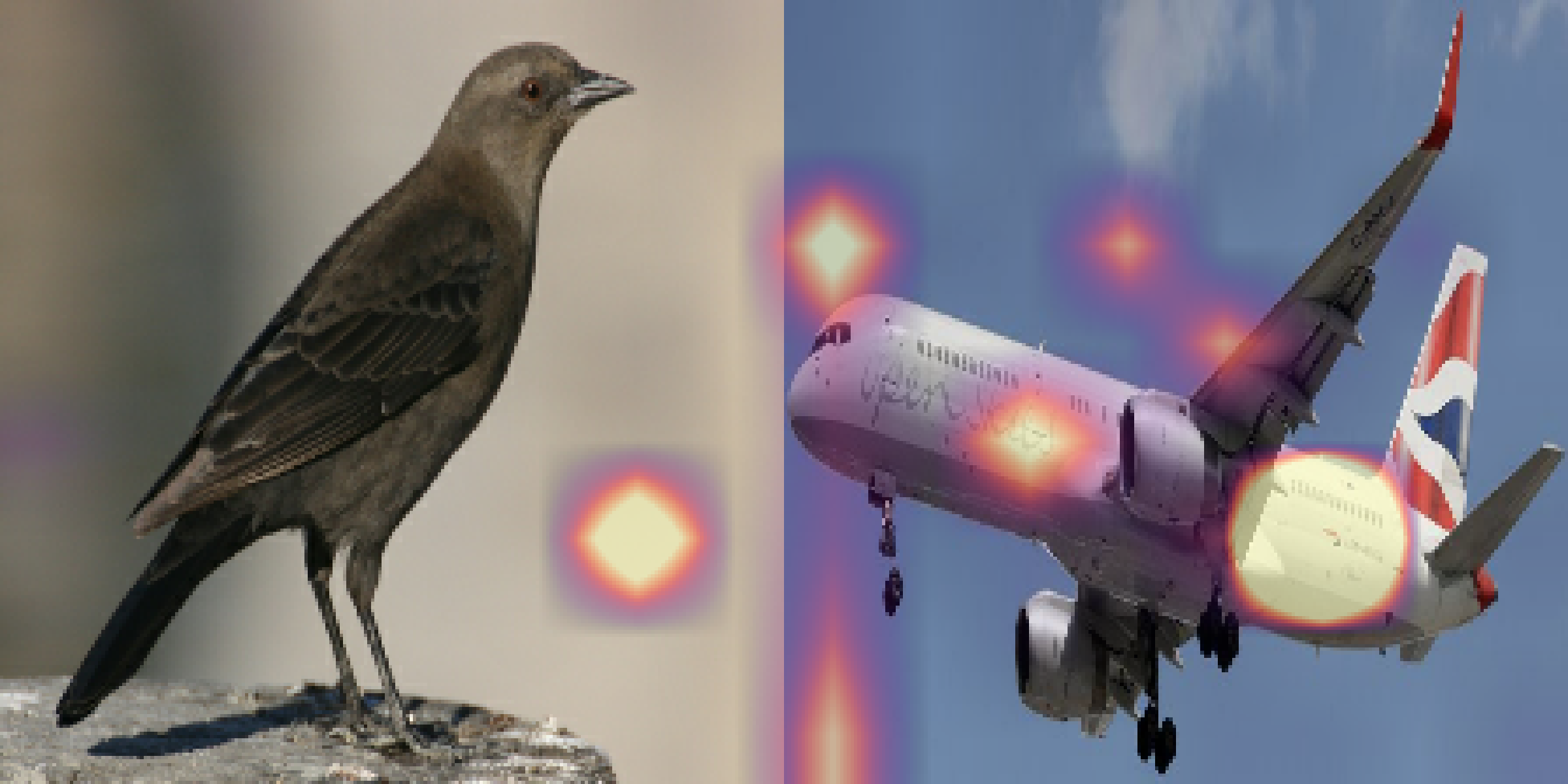} \\
    \vspace{-4pt} & \vspace{-4pt} \\
    \includegraphics[width=0.35\textwidth]{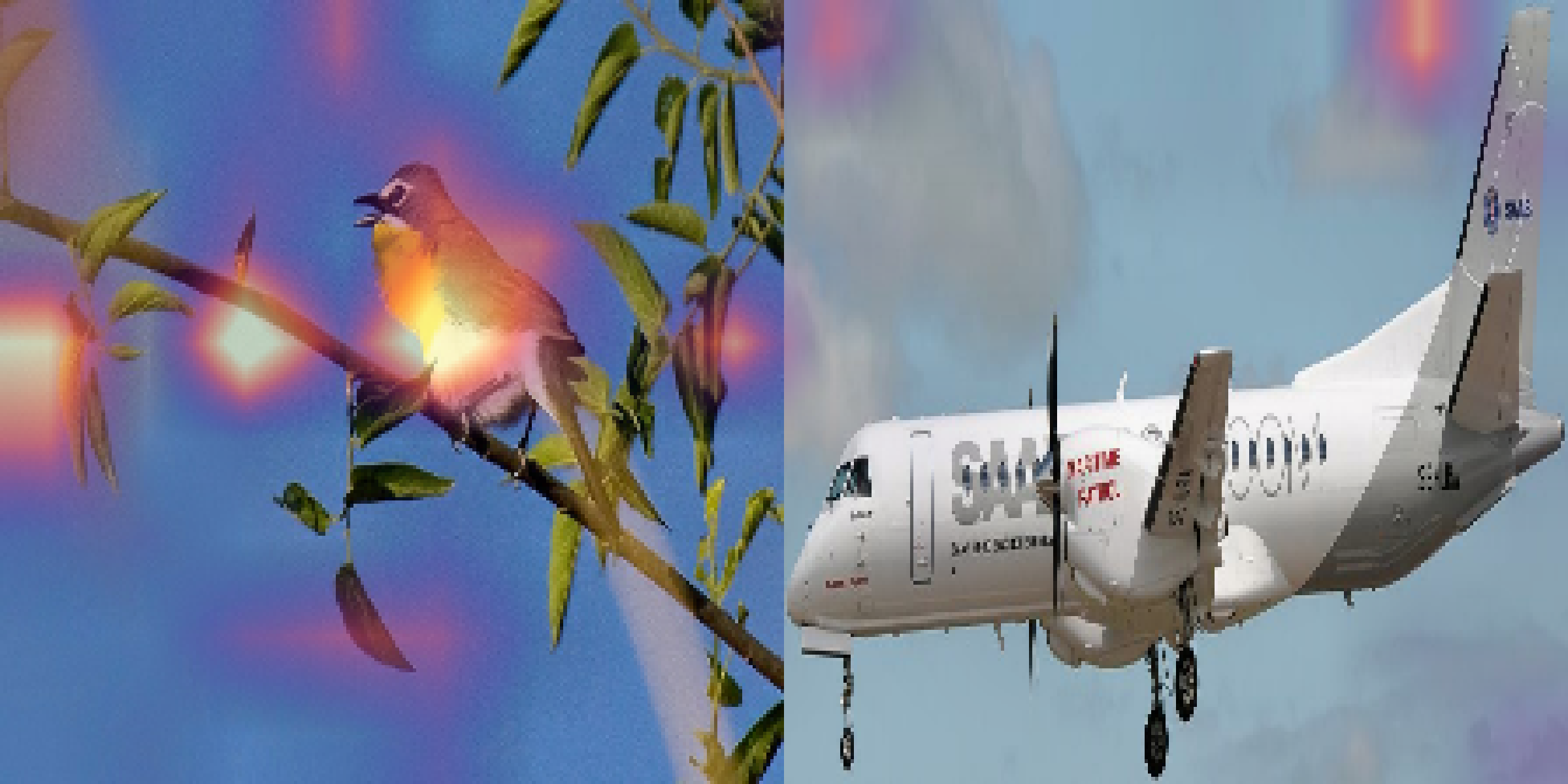} &
    \includegraphics[width=0.35\textwidth]{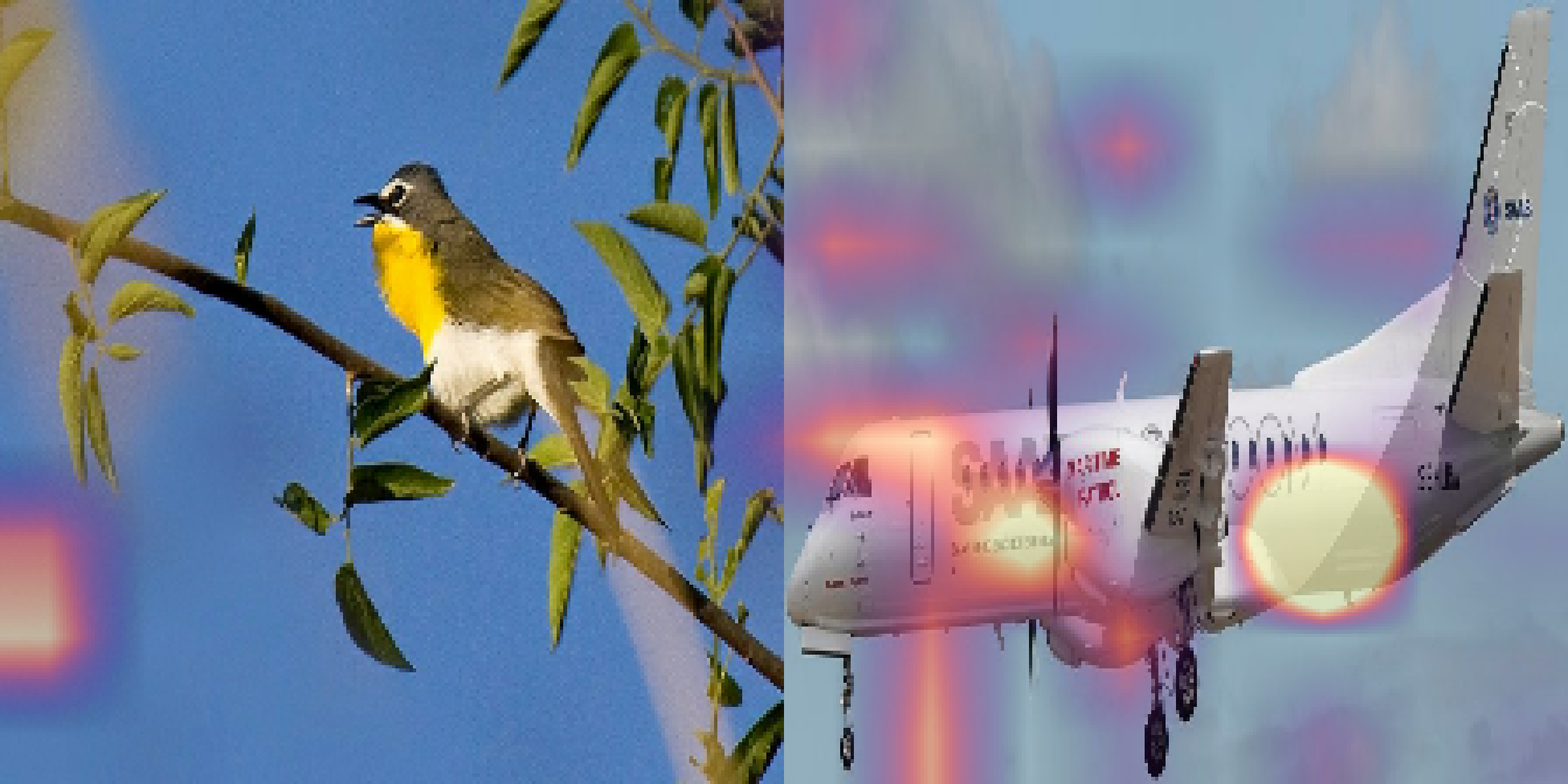} \\
    \vspace{-4pt} & \vspace{-4pt} \\
    \includegraphics[width=0.35\textwidth]{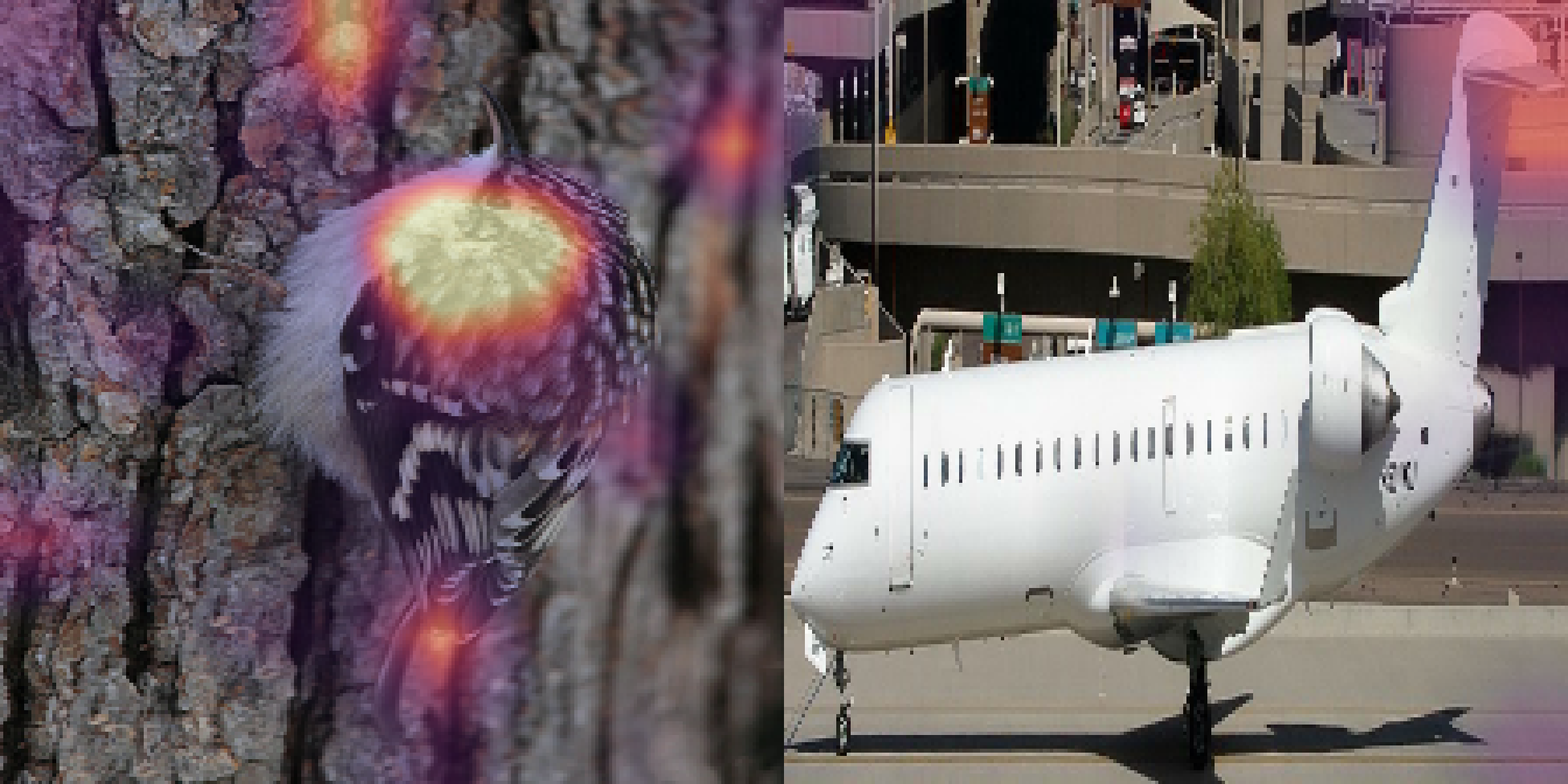} &
    \includegraphics[width=0.35\textwidth]{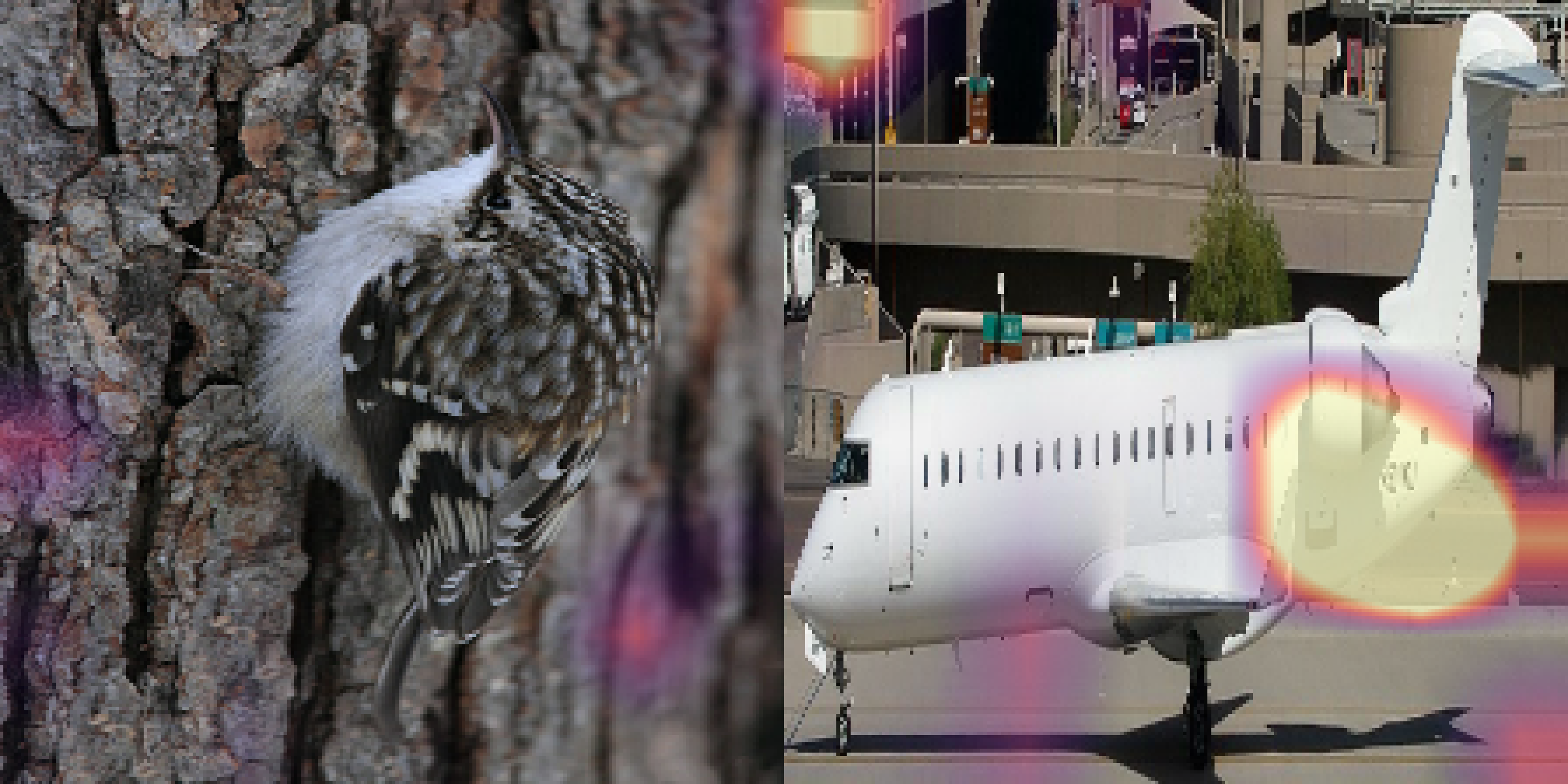} \\
    \\
    \vspace{-4pt} & \vspace{-4pt} \\
    {Prompt: \texttt{What type of bird is this?}} &
    {Prompt: \texttt{What type of plane is this?}}
  \end{tabular}
  \vspace{-4pt}
  \caption{
  \textbf{Top head attention map.}
  We concatenate bird~\cite{MD_BIRDS} and aircraft~\cite{CD_aircraft} datasets images horizontally in one support set. We then select the top vision-head for bird classification using the prompt \texttt{What type of bird is this?} and do the same for plane using the prompt \texttt{What type of plane is this?}. The attention map of the bird (left) and plane (right) top vision-head is overlaid on top of the image.}
  \label{fig:head-interpretability_2}
  \label{fig:interpretabiliy_2}
\end{figure}

Each accuracy is estimated on 1000 tasks. More precisely, we sample 100 10-way 4-shot tasks from each of the 10 standard datasets: EuroSAT~\cite{CD_eurosat}, UCF101~\cite{CD_ucf101}, DTD~\cite{CD_DTD}, Caltech101~\cite{CD_calthech_101}, SUN397~\cite{CD_sun}, OxfordPets~\cite{CD_oxford_pets}, StanfordCars~\cite{CD_stanford_cars}, Flowers102~\cite{CD_flowers_102}, Food101~\cite{CD_food_101}, and FGVCAircraft~\cite{CD_aircraft}.
The linear classifier used on each support set is a ridge classifier with regularization parameter $\lambda=1$, applied after L2 normalization of each vector.
We evaluate the accuracy on a query set composed of 5 examples per class (50 images in total).
On each figure, one tenth of the standard deviation of the accuracy across all tasks is shown as a color spread. As accuracy varies substantially from one task to another and from one dataset to another, we divide the standard deviation by 10 to improve the clarity of the figure. We believe that showing the standard deviation helps better understand how the figure is constructed.
\Cref{fig:interpretabiliy_2} shows 4 more examples of top-head attention maps. 

\begin{figure}[t]
\centering
\includegraphics[width=0.7\linewidth]{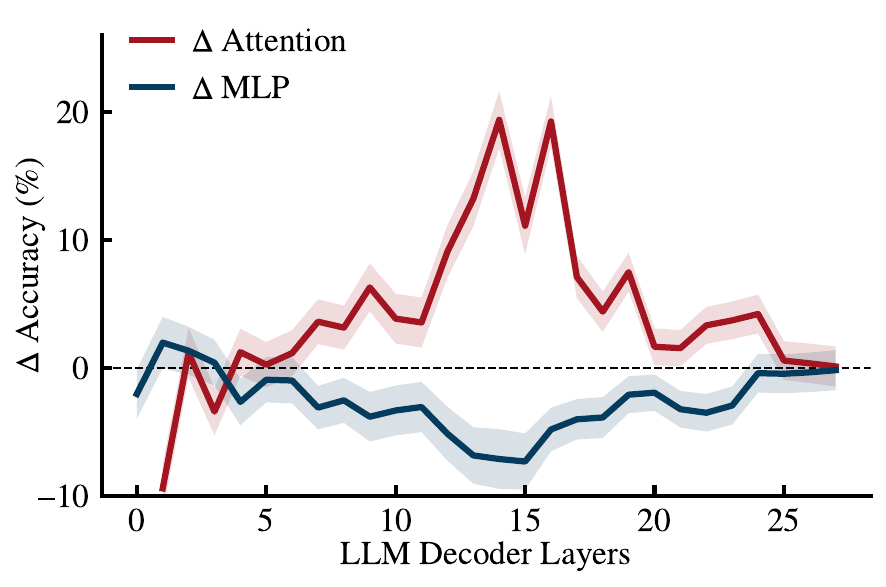}
\captionsetup{width=.92\linewidth}
\caption{Variation of accuracy after each Attention and MLP block.}
\label{fig:qwen_probing_b}
\end{figure}

We conduct an additional experiment to show the role of LVLM last token attention in building class discriminative multimodal representations.
More specifically, \cref{fig:qwen_probing_b} shows per-layer accuracy gain for MLP and attention blocks of the last token. Only attention blocks show positive gain across all layers, indicating that the last token improves representations by attending to both the text prompt and the vision tokens, with some layers contributing more than others.
 
\subsection{Evaluation Protocol}\label{sec:supp_eval}

For all methods, evaluation is conducted on the same set of seeds. All images are resized to $224 \times 224$, without using data augmentation strategies.

For the text-zero-shot setting, results are averaged over 100 tasks for each dataset, since evaluation is performed on a limited number of classes. 
To select the HEC-T top heads, we randomly sample 100 tasks on ImageNet and use the average query set accuracy of each head model as a ranking score.

For the vision-few-shot and text-vision-few-shot settings, hyperparameter search is required for each dataset. 
For each method, we select hyperparameters once using a single randomly sampled task (episode). 
We run the method over the sweep grid taken from the original paper and pick the configuration that maximizes query-set accuracy on that task (episode). 
We then fix this configuration for all remaining tasks (episodes) and report the resulting average performance. Results are averaged over 5 independently sampled tasks (episodes).
To select the HEC-T top heads, we use the support set as introduced in the method.

\paragraph{Models}
We describe below the implementation details for each model used. For CLIP-based and vision models, we use ViT-Base architecture, following SAVs~\cite{SAV}. Most implementations rely on either \texttt{Transformers}~\cite{wolf-etal-2020-transformers} or \\ \texttt{open\_clip}~\cite{cherti2023reproducible}:

\noindent\textbf{Vision models.}
\begin{itemize}
    \item \textbf{DINOv1}: implementation from \texttt{https://github.com/rashindrie/DIPA}.
    \item \textbf{DINOv2}: \texttt{Transformers}, repo ID \texttt{facebook/dinov2-base}.
    \item \textbf{DINOv3}: \texttt{Transformers}, repo ID\\ \texttt{facebook/dinov3-vitb16-pretrain-lvd1689m}.
\end{itemize}

\noindent\textbf{CLIP-based models.}
\begin{itemize}
    \item \textbf{SigLIP}: \texttt{Transformers}, repo ID \texttt{google/siglip-base-patch16-224}.
    \item \textbf{CLIP}: implementation from \texttt{https://github.com/mrflogs/ICLR24}.
    \item \textbf{DFN}: \texttt{open\_clip} implementation.
    \item \textbf{OpenCLIP}: \texttt{open\_clip} implementation.
\end{itemize}

\noindent\textbf{LVLM-based models.}
\begin{itemize}
    \item \textbf{Qwen2-VL}: \texttt{Transformers}, repo ID \texttt{Qwen/Qwen2-VL-7B-Instruct}.
    \item \textbf{LLaVA-OV}: \texttt{Transformers}, repo ID\\ \texttt{llava-hf/llava-onevision-qwen2-7b-ov-hf}.
    \item \textbf{Idefics2}: \texttt{Transformers}, repo ID \texttt{HuggingFaceM4/idefics2-8b}.
    \item \textbf{Finedefics}: \texttt{Transformers}, repo ID \texttt{StevenHH2000/Finedefics}.
\end{itemize}

\paragraph{Methods}
For baselines, we reuse the authors' public codebases and only modify the code required to interface them with our unified experimental framework. The corresponding repositories are:
\begin{itemize}
    \item \textbf{GDA}~\cite{CLIP_GDA}: \texttt{https://github.com/mrflogs/ICLR24}.
    \item \textbf{ProKeR}~\cite{CLIP_Proker}: \texttt{https://github.com/ybendou/ProKeR}.
    \item \textbf{Tip-Adapter}~\cite{CLIP_tipadapter}: \texttt{https://github.com/gaopengcuhk/Tip-Adapter}.
\end{itemize}


\subsection{Prompts}\label{sec:supp_prompts}

We provide prompts used for each domain dataset as well as general prompts used for task conditioning and no conditioning.
Each \textcolor{domain_color}{Domain} prompt was generated using Chat-GPT-5.2~\cite{openai_gpt52_chat_2025} and is shown in \cref{tab:domain-prompts}.
To measure the effect of task conditioning, we use the prompt \texttt{“Describe this image.”} as the prompt without conditioning (None) and \texttt{“What is on that image?”} as the prompt for \textcolor{task_color}{Task} conditioning. Because some datasets are not object-centric, a task prompt such as \texttt{"What object is in the image?"} is not general enough. 
For \textcolor{attn_color}{Class} conditioning, we append the candidate class texts to the prompt, one class per line giving $[\,\pi;\, \texttt{"\textbackslash n \{$t_{c_1}$\}"} ;\, \dots;\,  \texttt{"\textbackslash n \{$t_{c_N}$\}"}].$

\begin{table}[t]
\centering
\caption{Prompts used.}
\small
\begin{tabular}{ll}
\toprule
Dataset & \textcolor{domain_color}{Domain} prompt \\
\midrule
PETS & \texttt{What breed is the animal in this image?} \\
ESAT & \texttt{What type of remote sensing image does the given image belong to?} \\
UCF & \texttt{What action is the person performing in this video frame?} \\
SUN & \texttt{What scene is shown in this image?} \\
CAL & \texttt{What is the main object in this photo?} \\
DTD & \texttt{What texture pattern is visible in this image?} \\
AIR & \texttt{Name the aircraft model shown.} \\
FOOD & \texttt{What is this dish called?} \\
FLWR & \texttt{What is the species of this flower?} \\
CARS & \texttt{Which car model is shown in the image?} \\
BIRD & \texttt{What is the species of this bird?} \\
SIGN & \texttt{What is the type of this traffic sign?} \\
\midrule
\midrule
- & Other prompt conditioning \\
\midrule
None & \texttt{Describe this image.} \\
\textcolor{task_color}{Task} & \texttt{What is on that image?} \\
\bottomrule
\end{tabular}
\label{tab:domain-prompts}
\end{table}

\subsection{Computational Cost}
\label{sec:supp_cost}

We compare the computational cost of HEC-V to that of linear probing, focusing only on the classifier fitting step once support set features have been extracted. In Qwen2-VL, the LLM decoder has \(L=28\) layers and \(H=28\) attention heads per layer, and the hidden size is 3584, which gives a per-head dimension of \(D=128\). A ridge linear probe fitted on the summary-token therefore operates on features of dimension \(D H\), so, in the closed-form formulation, its dominant cost is the inversion of a \((D H)\times(D H)\) regularized covariance matrix, yielding a complexity of \(\mathcal{O}((D H)^3)\). By contrast, HEC-V fits one Gaussian model per head and inverts \(L H\) covariance matrices of size \(D\times D\), which yields a total complexity of \(\mathcal{O}(L H D^3)\). Hence, the classifier fitting stage of HEC-V is more efficient than linear probing by a factor
\begin{equation}
\frac{(D H)^3}{L H D^3} = \frac{H^2}{L}.
\end{equation}
For Qwen2-VL, this corresponds to a factor of \(28\). 

\subsection{Expression of the Constant $C$}
\label{sec:supp_constant_C}

In Eq.~(6) of the main paper, the class logit is written as
\begin{equation}
\ell_{i,m,c}
=
-\frac{1}{2}
\Big(
h^{(v)}_{i,m}-\hat{\mu}_{m,c}
\Big)^{\top}
\hat{\Sigma}_m^{-1}
\Big(
h^{(v)}_{i,m}-\hat{\mu}_{m,c}
\Big)
+ C.
\end{equation}

The constant $C$ groups all terms that do not depend on the class index $c$.
Starting from the Gaussian discriminant model, we have
\begin{equation}
\log p\!\left(h^{(v)}_{i,m}, y=c\right)
=
\log p\!\left(h^{(v)}_{i,m}\mid y=c\right)
+
\log p(y=c),
\end{equation}
and, since
\begin{equation}
\begin{aligned}
\log p\!\left(h^{(v)}_{i,m}\mid y=c\right)
&=
-\frac{1}{2}
\Big(
h^{(v)}_{i,m}-\hat{\mu}_{m,c}
\Big)^{\top}
\hat{\Sigma}_m^{-1}
\Big(
h^{(v)}_{i,m}-\hat{\mu}_{m,c}
\Big)
\\
&\quad
-\frac{1}{2}\log \left|\hat{\Sigma}_m\right|
-\frac{D}{2}\log(2\pi),
\end{aligned}
\end{equation}
it follows that
\begin{equation}
C
=
-\frac{1}{2}\log \left|\hat{\Sigma}_m\right|
-\frac{D}{2}\log(2\pi)
+
\log p(y=c).
\end{equation}

In our episodic $N$-way $K$-shot setting, each class is sampled with the same number of support examples, so we use a uniform class prior
\begin{equation}
p(y=c)=\frac{1}{N}.
\end{equation}
Therefore,
\begin{equation}
C
=
-\frac{1}{2}\log \left|\hat{\Sigma}_m\right|
-\frac{D}{2}\log(2\pi)
-\log N,
\end{equation}
which is independent of $c$. As a consequence, $C$ cancels out in the softmax used to compute class probabilities, and also does not affect the $\arg\max_c$ prediction rule.

\section{Ablations}\label{sec:supp_ABLATION}


\begin{table}[t]
\centering
\caption{Comparison of ensemble methods.}
\setlength{\tabcolsep}{6pt}
\begin{tabular}{llc}
\hline
Category & Method & Acc. ($\%$) \\
\hline
Voting & Majority vote & $93.85_{0.41}$ \\
Voting & Weighted vote & $93.92_{0.41}$ \\
\hline
Proba & Mean & $94.00_{0.40}$ \\
Proba & Score weights & $94.01_{0.40}$ \\
Proba & Optimal weights & $\mathbf{94.04_{0.42}}$ \\
\hline
Logit & Mean & $93.94_{0.41}$ \\
Logit & Score weights & $93.95_{0.41}$ \\
Logit & Optimal weights & $93.98_{0.41}$ \\
\hline
\end{tabular}
\label{tab:ablation_ensemble}
\end{table}

\subsection{Ablation of Ensemble Methods}\label{sec:supp_ensemble}
We conduct a series of experiments to study how different head ensembling strategies affect performance.
For HEC-V, we evaluate 4-shot 10-way classification over 300 tasks across 10 datasets.
\Cref{tab:ablation_ensemble} reports the results.

We compare the following ensemble variants, all applied to the top-$k$ vision-heads $H_V$:

\begin{itemize}
    \item \textbf{Majority vote.}
    Each head predicts a label $\hat{y}_{q,m}$. The final prediction is
    \begin{equation}
    \hat{y}_q
    =
    \arg\max_c \sum_{m \in H_V} \mathbf{1}\!\left[\hat{y}_{q,m}=c\right],
    \end{equation}
    where $\mathbf{1}[\cdot]$ denotes the indicator function.

    \item \textbf{Weighted vote.}
    Same as majority vote, but each head vote is weighted by its ranking score $s_m^{(v)}$. The final prediction is
    \begin{equation}
    \hat{y}_q
    =
    \arg\max_c \sum_{m \in H_V} s_m^{(v)} \, \mathbf{1}\!\left[\hat{y}_{q,m}=c\right].
    \end{equation}

    \item \textbf{Logit Mean.}
    We average the logits
    \begin{equation}
    \bar{\ell}_{q,c}=\frac{1}{|H_V|}\sum_{m \in H_V}\ell_{q,m,c},
    \end{equation}
    and predict with $\arg\max_c \bar{\ell}_{q,c}$.

    \item \textbf{Logit Score weights.}
    We compute a weighted sum of logits
    \begin{equation}
    \bar{\ell}_{q,c}=\frac{\sum_{m \in H_V} s_m^{(v)} \ell_{q,m,c}}{\sum_{m \in H_V} s_m^{(v)}},
    \end{equation}
    and predict with $\arg\max_c \bar{\ell}_{q,c}$.

    \item \textbf{Logit Optimal weights.}
    We learn weights $\{w_m\}_{m \in H_V}$ on the support set by minimizing
    \begin{equation}
    \sum_i \left\| \mathbf{y}_i-\sum_{m \in H_V} w_m \, \boldsymbol{\ell}_{i,m} \right\|_2^2
    + \lambda \|\mathbf{w}\|_2^2,
    \end{equation}
    where $\mathbf{y}_i$ is the one-hot label vector, $\boldsymbol{\ell}_{i,m}$ is the logit vector predicted by head $m$, and we set $\lambda=1$.

    \item \textbf{Proba Mean.}
    We average the head class probabilities
    \begin{equation}
    \bar{p}_{q,c}=\frac{1}{|H_V|}\sum_{m \in H_V} p^{(v)}_{q,m,c},
    \end{equation}
    which corresponds to HEC-V in Eq.~(9) (main paper).

    \item \textbf{Proba Score weights.}
    We compute a weighted sum of probabilities
    \begin{equation}
    \bar{p}_{q,c}=\frac{\sum_{m \in H_V} s_m^{(v)} p^{(v)}_{q,m,c}}{\sum_{m \in H_V} s_m^{(v)}}.
    \end{equation}

    \item \textbf{Proba Optimal weights.}
    We learn weights $\{w_m\}_{m \in H_V}$ on the support set by minimizing
    \begin{equation}
    \sum_i \left\| \mathbf{y}_i-\sum_{m \in H_V} w_m \, \mathbf{p}^{(v)}_{i,m} \right\|_2^2
    + \lambda \|\mathbf{w}\|_2^2,
    \end{equation}
    where $\mathbf{y}_i$ is the one-hot label vector, $\mathbf{p}^{(v)}_{i,m}$ is the class probability vector predicted by head $m$, and we set $\lambda=1$.
\end{itemize}

Ensembling probabilities performs best overall, although all methods give similar results.
\textbf{Voting} remains competitive despite its simplicity.
We use the \textbf{Proba Mean} formulation for HEC because it is simple, robust, and does not introduce additional hyperparameters.

\begin{figure}[t]
\centering
\includegraphics[width=0.6\linewidth]{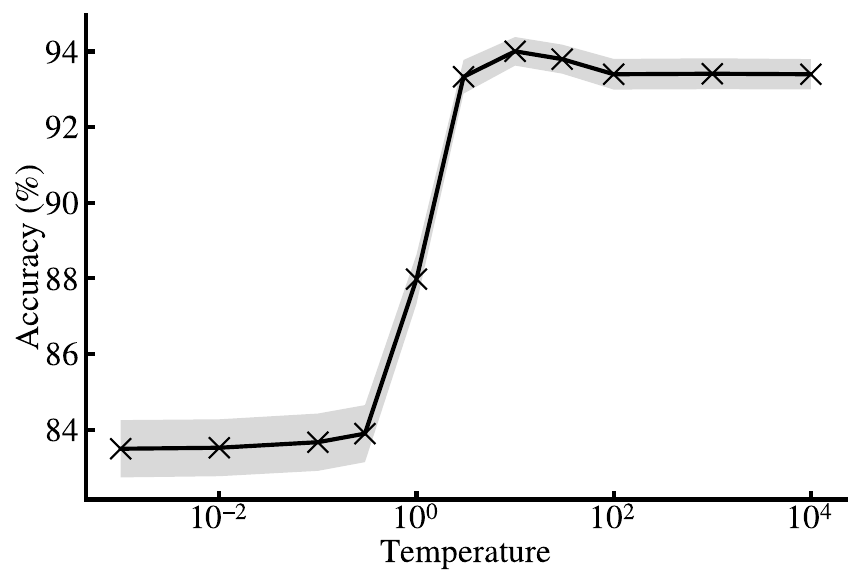}
\captionsetup{width=.92\linewidth}
\caption{Ablation Study of the hyperparameter $\tau$}
\label{fig:tau}
\end{figure}

\subsection{Ablation of the Temperature $\tau$}\label{sec:supp_tau}

We study the impact of the temperature hyperparameter $\tau$ on HEC-V over 300 10-way 4-shot tasks across 10 datasets. The results are shown in \cref{fig:tau}. $\tau=10$ performs best in this setting.
More generally, higher values outperform lower ones. As explained in the method section, this comes from avoiding the saturation of the support set accuracy when ranking heads.
Thus, we advise using higher values of $\tau$ for smaller support sets with an increased chance of overfitting.


\subsection{Ablation of Head Selection}\label{sec:supp_heads}

Similarly to prompts, which can be specific to a task, a domain, or a set of classes, we conduct a series of experiments to assess whether heads are specific to a task, a domain, or a set of classes. We therefore also assess the transferability of top heads from one task to another.

First, we select the top \textcolor{task_color}{Task} heads on ImageNet, by ranking heads according to their average query set accuracy over 100 10-way 4-shot tasks on ImageNet.
Then, we select the top \textcolor{domain_color}{Domain} heads on their respective domain datasets, by ranking with the best average query set accuracy over 100 10-way 4-shot tasks.
The ranking score is called HEC Oracle as we use the query set accuracy as a ranking score.

Finally, we use our method to rank, from the support set at test time, the best \textcolor{attn_color}{Class} head for any given 10-way 4-shot task. It is important to note that ranking heads from the support set is harder, and only provides a proxy for query set accuracy.
We call that head ranking method HEC Test-time.

Results are shown in \cref{tab:head_context}. 
We see that text-heads are shared across class, domain, and task. 
Selecting on the fly from the support set the best heads is comparable to knowing in advance the best-performing heads for a given domain. More precisely, we observe a small performance drop when ranking heads from the support set in that setup.

Vision-heads are less transferable, as the domain heads perform on average 0.4\% better than general \textcolor{task_color}{Task} heads. Similarly to text-heads, selecting the best domain heads in advance performs slightly better than selecting at test time for a given task in the 10-way 4-shot setup.

\begin{figure}[t]
    \centering
    \begin{subfigure}[t]{0.49\linewidth}
        \centering
        \includegraphics[width=\linewidth]{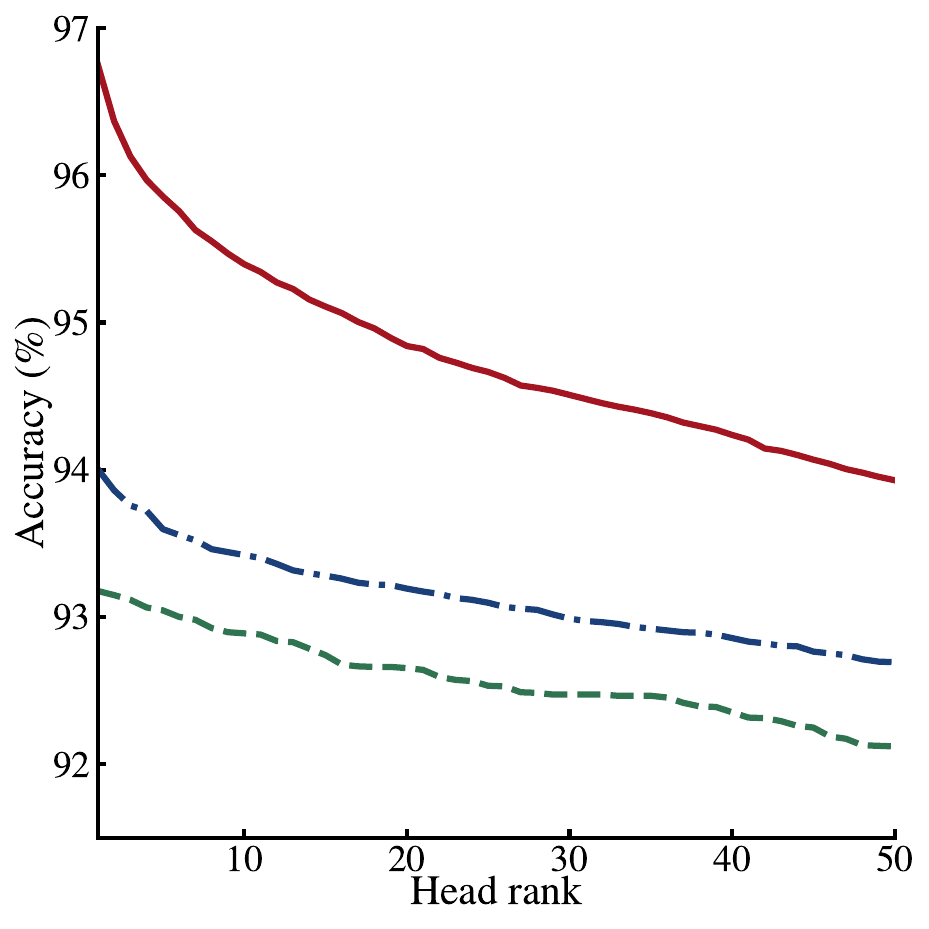}
        \caption{Vision-head rank}
        \label{fig:head_rank_qwen_v}
    \end{subfigure}\hfill
    \begin{subfigure}[t]{0.49\linewidth}
        \centering
        \includegraphics[width=\linewidth]{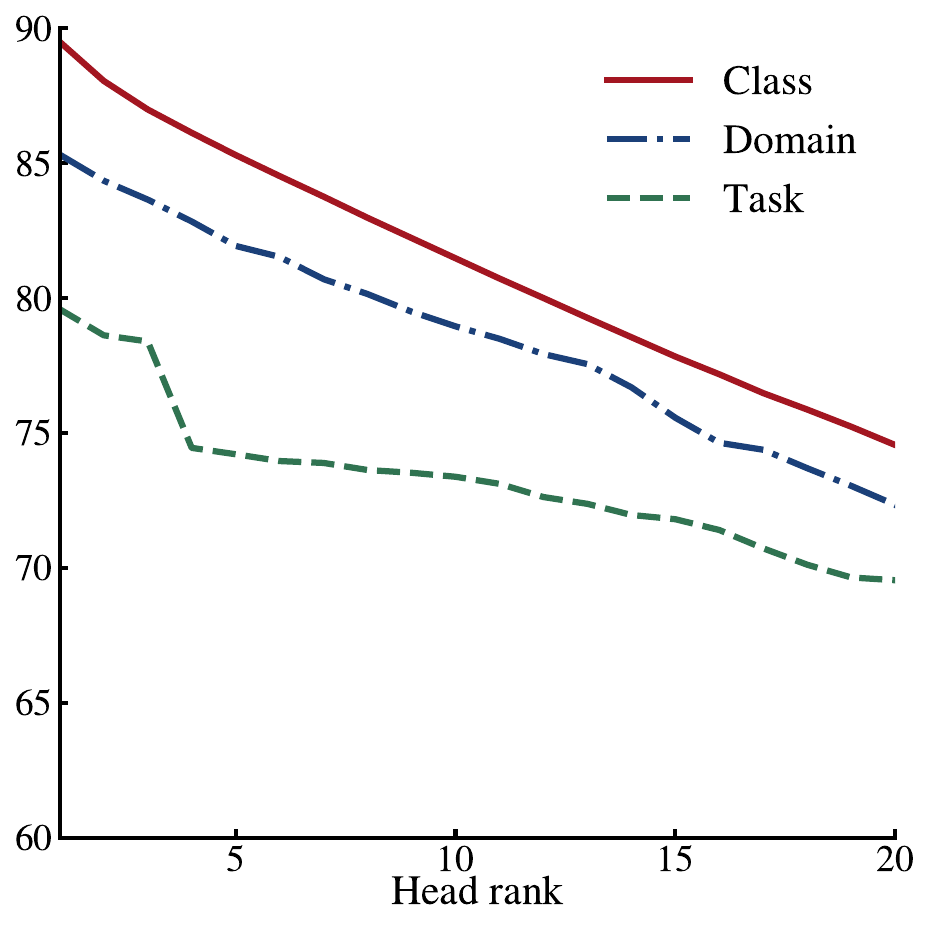}
        \caption{Text-head rank}
        \label{fig:head_rank_qwen_t}
    \end{subfigure}
    \caption{Average accuracy of the top 50 heads. 
    \textcolor{task_color}{Task} heads are ranked on ImageNet query set accuracy. 
    \textcolor{domain_color}{Domain} heads are ranked on the domain dataset query set accuracy.
    \textcolor{attn_color}{Class} heads are ranked on the 10-way 4-shot current task query set accuracy.
    }
    \label{fig:head_rank_qwen}
\end{figure}

\Cref{fig:head_rank_qwen} shows, for both text-heads and vision-heads, the average accuracy of the top 50 \textcolor{attn_color}{Class}, \textcolor{domain_color}{Domain}, and \textcolor{task_color}{Task} heads. In particular, we observe that three text-heads in Qwen2-VL stand out and consistently achieve notably higher zero-shot accuracy than the others.

\begin{figure}[t]
    \centering
    \begin{subfigure}[t]{0.49\linewidth}
        \centering
        \includegraphics[width=\linewidth]{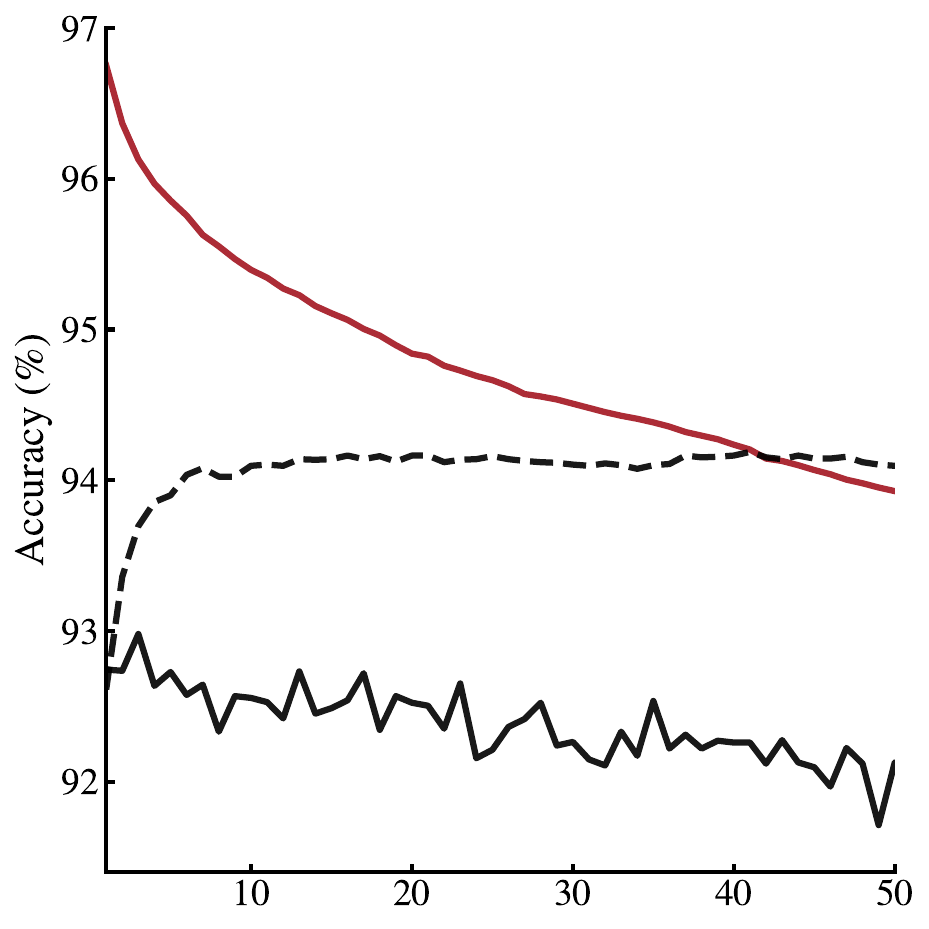}
        \caption{Vision-head ensemble}
        \label{fig:head_hec_qwen_v}
    \end{subfigure}\hfill
    \begin{subfigure}[t]{0.49\linewidth}
        \centering
        \includegraphics[width=\linewidth]{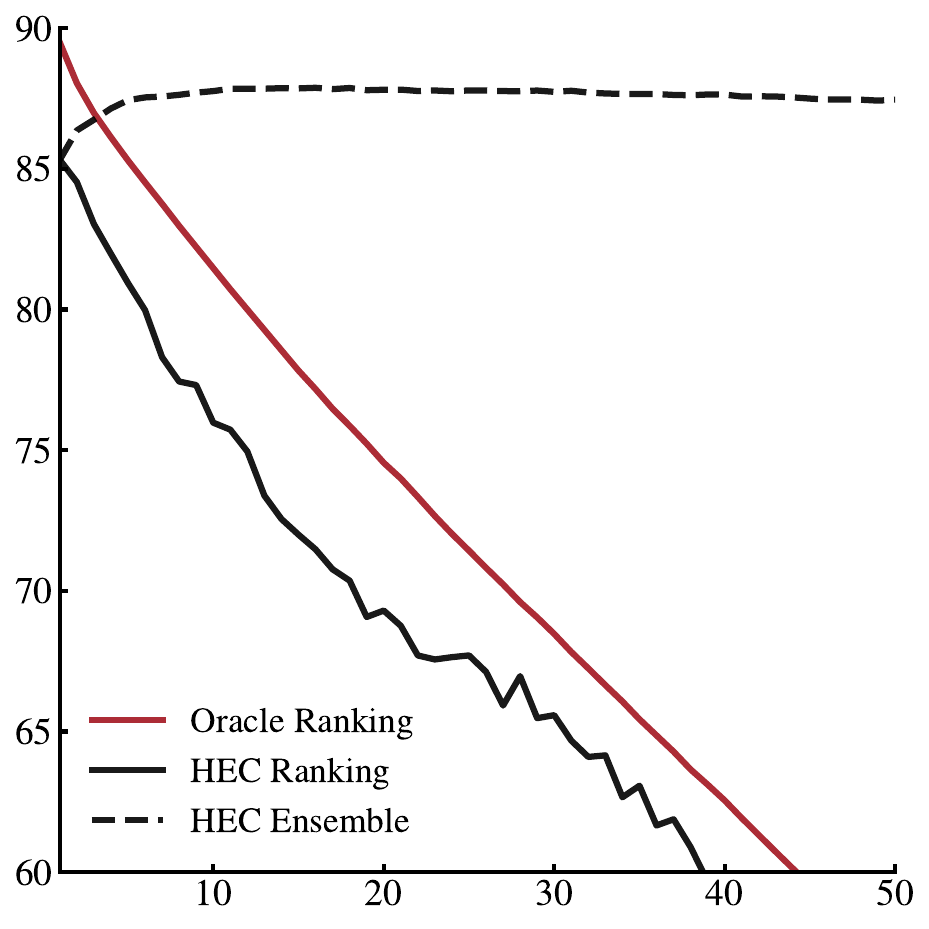}
        \caption{Text-head ensemble}
        \label{fig:head_hec_qwen_t}
    \end{subfigure}
    \caption{Enlarged versions of Figs. 5b and 5c from the main paper.}
    \label{fig:head_hec_qwen}
\end{figure}

\Cref{fig:head_hec_qwen} is an enlarged version of Figs. 5b and 5c from the main paper. 
This figure shows the gap between head ranking with HEC on the support set and an oracle ranking based on query set accuracy. It also shows the effect of varying the number top-$k$ of heads included in the ensemble.
We observe that ensembling is robust to the choice of top-$k$ for HEC-V and HEC-T. In particular, aggregating the top 10 heads yields a strong improvement. Beyond that point, adding less discriminative heads does not lead to a decrease in accuracy, especially for vision-heads, where adding more heads further improves performance.

\begin{table}[t]
    \centering  
    \caption{
Head Selection. 
We evaluate HEC using different sets of heads.
\textcolor{task_color}{Task} heads are the top 20 on ImageNet. 
\textcolor{domain_color}{Domain} heads are the top 20 on the given dataset, 
and \textcolor{attn_color}{Class} heads are selected on the fly on a given support set by our method HEC, without knowing in advance the performance on the query set.
}
    {%
    \setlength{\tabcolsep}{6pt}
    \begin{tabular}{lll|ll|ll}
    \toprule
    \multicolumn{3}{c}{} & \multicolumn{2}{c}{HEC-T} & \multicolumn{2}{c}{HEC-V} \\
    \cmidrule(lr){4-5}\cmidrule(lr){6-7}
    \cmidrule(lr){4-6}
    Heads & Selection Method &  & Acc. & Gain & Acc. & Gain \\
    \midrule
    \textcolor{task_color}{Task}   & HEC Oracle    &  & $88.52_{0.6}$ &  & $93.79_{0.4}$ &  \\
    \textcolor{domain_color}{Domain} & HEC Oracle  &  & $88.53_{0.6}$ & $\textcolor{darkgreen}{\uparrow +\mkern1mu0.01}$ & $94.18_{0.4}$ & $\textcolor{darkgreen}{\uparrow +\mkern1mu0.39}$ \\
    \textcolor{attn_color}{Class} & HEC Test-time &  & $88.45_{0.6}$ & $\textcolor{darkred}{\downarrow -\mkern1mu0.08}$ & $94.14_{0.4}$ & $\textcolor{darkred}{\downarrow -\mkern1mu0.04}$ \\
    \bottomrule
    \end{tabular}
    }%
    \label{tab:head_context}
\end{table}

\subsection{Failing Case of Class Conditioning}\label{sec:supp_failling}

In this section, we study how class conditioning is affected by the number of classes $N$. For this experiment, we evaluate the performance of HEC-T with \textcolor{domain_color}{Domain} and \textcolor{attn_color}{Class} conditioning. 
We additionally compare against the letter-prompt zero-shot baseline up to $N=25$, as we are limited by the number of letters in the alphabet.
We evaluate HEC-T on varying $N$-way tasks, reporting the average accuracy over 300 tasks across 10 datasets.
When a dataset does not contain enough classes for a given $N$, we use the maximum available number of classes. Note that EuroSAT has only 10 classes.

\begin{figure}[t]
\centering
\includegraphics[width=0.6\linewidth]{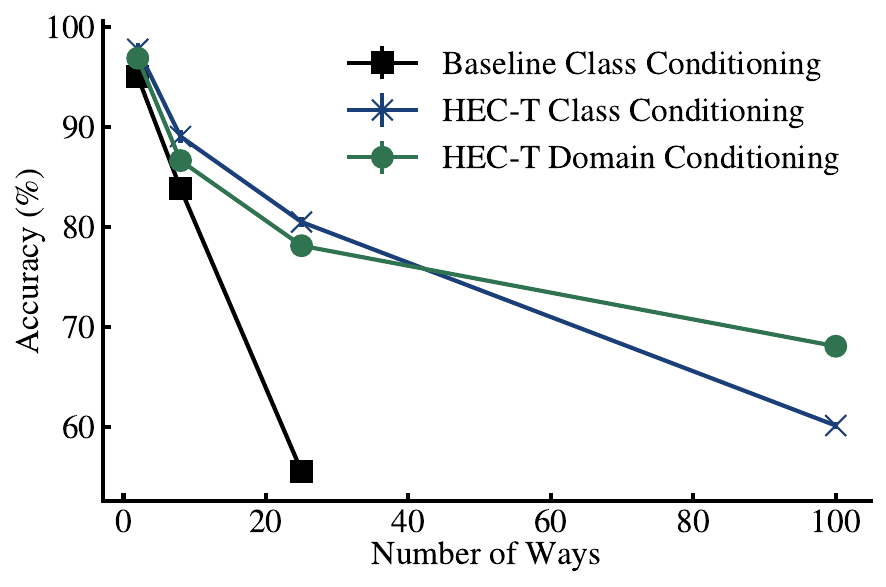}
\captionsetup{width=.92\linewidth}
\caption{Impact of $N$-ways on \textcolor{attn_color}{Class} and \textcolor{domain_color}{Domain} Conditioning Performance on HEC-T}
\label{fig:nways}
\end{figure}

\Cref{fig:nways} shows that, for a small number of classes, below 25, class conditioning performs better. However, when the number of classes increases to 100, domain conditioning performs better. This indicates that including too many classes in the prompt eventually leads to a degradation, highlighting one of the limits of our method. It also shows that, when evaluating with a large number of classes, domain conditioning is preferred. 
We observe that the performance gap between the baseline and HEC-T increases as $N$ grows.

\section{Additional Experiment Results}\label{sec:supp_RESULT}

\subsection{Image-Text Classification}\label{sec:supp_itc}
In this section we show the performance of our method on an image-text classification task. More specifically, we evaluate on the Image-Text Retrieval benchmark NaturalBench-Retrieval~\cite{li2024naturalbench}.
It consists in determining whether a given image-caption pair corresponds.
Each image-text pair is assigned a binary label: Yes if they match, and No otherwise.
NaturalBench-Retrieval is made challenging by using two similar images with two corresponding captions, effectively eliminating language bias and requiring models to capture more nuanced visual-semantic relationships.
Following the benchmark procedure, we evaluate 
text accuracy (T) (when the model correctly answers both questions for a text), 
image accuracy (I) (when the model correctly answers both questions for an image)
, and
group accuracy (G) (when the model correctly answers all four pairs).
We follow the 2-way 20-shot evaluation setup of SAVs~\cite{SAV}, and report the results from the paper.
We evaluate HEC-V by adding our implementation to the SAVs codebase.
We compare our approach against several state-of-the-art baselines, including closed-sourced GPT-4o \cite{openai2024gpt4ocard}, open vision language models LLaVA-1.5~\cite{liu2023llava15} and Instruct-BLIP~\cite{dai2023instructblip}. Zero-shot baselines are obtained by prompting each model directly and decoding an answer. We also compare against few-shot test-time adaptation and finetuning approaches, including MTV~\cite{MTV}, SAVs~\cite{SAV}, as well as 4-shot in-context learning and LoRA finetuning \cite{hu2022lora}. 
Results are shown in \cref{tab:results_main}.

\input{table/sota_sav}

\subsection{Other Models}\label{sec:supp_other_models}


In this section, we study whether the head selection mechanism of HEC-VT transfers to other models. 
We use exactly the same setup as in the text-vision-few-shot setting and evaluate LLaVA-OV, as well as another open-source LVLM, Idefics2~\cite{laurencon2024matters}. 
More interestingly, we also evaluate Finedefics~\cite{Finedefics}, a finetuned version of Idefics2 specifically trained for fine-grained image classification, to verify that our method is complementary to finetuning.
Results are reported in \Cref{tab:model-baselines} against summary-token linear probing (Probing) and summary-token zero-shot (Zero-Shot). 
HEC-VT improves performance over Probing for all three models, by 2.1, 3.4, and 2.9 points on Finedefics, Idefics2, and LLaVA-OV, respectively. 
Finedefics indeed has stronger zero-shot performance compared to Idefics2 (+12.7\%).
HEC-VT further improves its text-vision-few-shot performance, yielding the best overall result of 84.6\%.

\begin{table}[t]
  \centering
  \caption{\textbf{text-vision-few-shot} average accuracy (\%) across models. \textbf{Bold} denotes the best method for each model. \textcolor{darkgreen}{Green} denotes the absolute gain of HEC-VT over Probing.}
  \vspace{-.4em}
  \setlength{\tabcolsep}{12pt}
  \begin{tabular}{llll}
    \toprule
    Method & Finedefics~\cite{Finedefics} & Idefics2~\cite{laurencon2024matters} & LLaVA-OV~\cite{li2024llavaOV} \\
    \midrule
    Zero-Shot & $56.6$ & $43.9$ & $41.6$ \\
    Probing & $82.5$ & $80.3$ & $81.6$ \\
    \hec{HEC-VT} & \hec{$\mathbf{84.6}$} & \hec{$\mathbf{83.6}$} & \hec{$\mathbf{84.5}$} \\
     & \textcolor{darkgreen}{\raisebox{0.2ex}{\scriptsize +}2.1} & \textcolor{darkgreen}{\raisebox{0.2ex}{\scriptsize +}3.4} & \textcolor{darkgreen}{\raisebox{0.2ex}{\scriptsize +}2.9} \\
    \bottomrule
  \end{tabular}
  \label{tab:model-baselines}
\end{table}



%% file: table/sota_sav.tex
\begin{table*}[t]
\centering
\scriptsize
\setlength{\tabcolsep}{4pt}
\renewcommand{\arraystretch}{1.12}
\newcommand{\best}[1]{\textbf{#1}}
\newcommand{\second}[1]{\underline{#1}}
\newcommand{\na}{--}

\caption{Results on Image-Text Retrieval benchmark. Best is shown in \best{bold}. Baselines are shaded in blue.}
\begin{adjustbox}{max width=\textwidth}
\begin{tabular}{lccc}
\toprule
& \multicolumn{3}{c}{NaturalBench Retrieval} \\
\cmidrule(lr){2-4}
Model
& T & I & G \\
\midrule

\rowcolor{blue!15}CLIP & 41.8 & 45.0 & 23.2 \\
\rowcolor{blue!15}SigLip & 54.5 & 54.9 & 31.2 \\
\rowcolor{blue!15}GPT-4o & 65.0 & 67.0 & 40.5 \\
\rowcolor{blue!15}LLaVA-1.5 & 36.7 & 42.7 & 12.2 \\
\rowcolor{blue!15}Instruct-BLIP & 19.5 & 21.3 & 1.1 \\
\midrule

Qwen2-VL & 60.2 & 61.9 & 35.6 \\
\hspace{1em}+4-shot-ICL & 42.4 & 45.6 & 22.7 \\
\hspace{1em}+MTV~\cite{MTV} & 63.5 & 64.0 & 37.0 \\
\hspace{1em}+LoRA & 65.2 & 66.1 & 40.4 \\
\hspace{1em}+SAVs~\cite{SAV} & 70.0 & 71.0 & 42.5 \\
\arrayrulecolor{gray!55}
\cmidrule[0.3pt](l{6pt}r{2pt}){2-4}
\arrayrulecolor{black}
\rowcolor{gray!15} \hspace{1em}+HEC-V (Ours) & \best{71.9} & \best{73.0} & \best{51.9} \\
\bottomrule
\end{tabular}
\end{adjustbox}

\label{tab:results_main}
\end{table*}